%% file: main.tex
\documentclass[10pt,twocolumn,letterpaper]{article}

\usepackage[pagenumbers]{cvpr} 

\input{preamble}

\newcommand{\parag}[1]{\vskip4pt \noindent \textbf{#1}}
\newcommand{\boldparagraph}[1]{\vspace{0.0em}\noindent{\bf #1} }

\makeatletter
\@namedef{ver@everyshi.sty}{}
\makeatother

\usepackage{tikz}
\usepackage{balance}
\usepackage{multirow}
\usepackage{balance}
\newcommand{\name}{ALSTER}
\usepackage{arydshln}
\usepackage{balance}

\definecolor{cvprblue}{rgb}{0.21,0.49,0.74}
\usepackage[pagebackref,breaklinks,colorlinks,citecolor=cvprblue]{hyperref}

\def\paperID{5} 
\def\confName{3DV\xspace}
\def\confYear{2024\xspace}

\title{ALSTER: \underline{A} \underline{L}ocal \underline{S}patio-\underline{T}emporal \underline{E}xpert for \\Online 3D Semantic \underline{R}econstruction}

\author{
Silvan Weder$^{1}$ \qquad Francis Engelmann$^{1,4}$ \qquad Johannes L. Schönberger$^{2}$\qquad Akihito Seki$^{3}$
\vspace{5px}
\\
Marc Pollefeys$^{1, 2}$ \qquad Martin R. Oswald$^{1, 5}$ \vspace{5px}\\
{ 
$^{1}$ ETH Zurich \quad
$^{2}$ Microsoft \quad
$^{3}$ Toshiba \quad
$^{4}$ Google \quad
$^{5}$ University of Amsterdam\quad
}
}

\begin{document}
\maketitle
\input{sec/00_abstract}

\input{sec/01_introduction}
\input{sec/02_relatedwork}
\input{sec/03_method}
\input{sec/04_experiments}
\input{sec/05_conclusion}

\newpage
\small
\bibliographystyle{ieeenat_fullname}
\bibliography{egbib}

\end{document}


\maketitle

\begin{abstract}
In this supplementary material, we provide additional implementations details of our pipeline (Sec.~\ref{sec:pipeline}), we present additional results and discuss the compared baselines (Sec.~\ref{sec:results}).
\end{abstract}

\tableofcontents

\input{sec/06_appendix}


\clearpage
{\small
\bibliographystyle{ieee_fullname}
\bibliography{egbib}
}

%% file: preamble.tex
\usepackage[dvipsnames]{xcolor}
\usepackage{colortbl}


%% file: sec/00_abstract.tex
\begin{abstract}
We propose an online 3D semantic segmentation method that incrementally reconstructs a 3D semantic map from a stream of RGB-D frames.
Unlike offline methods, ours is directly applicable to scenarios with real-time constraints, such as robotics or mixed reality. 
To overcome the inherent challenges of online methods, we make two main contributions.
First, to effectively extract information from the input RGB-D video stream, we jointly estimate geometry and semantic labels per frame in 3D.
A key focus of our approach is to reason about semantic entities both in the 2D input and the local 3D domain to leverage differences in spatial context and network architectures.
Our method predicts 2D features using an off-the-shelf segmentation network.
The extracted 2D features are refined by a lightweight 3D network to enable reasoning about the local 3D structure. 
Second, to efficiently deal with an infinite stream of input RGB-D frames, a subsequent network serves as a temporal expert predicting the incremental scene updates by leveraging 2D, 3D, and past information in a learned manner. These updates are then integrated into a global scene representation.
Using these main contributions, our method can enable scenarios with real-time constraints and can scale to arbitrary scene sizes by processing and updating the scene only in a local region defined by the new measurement.
Our experiments demonstrate improved results compared to existing online methods that purely operate in local regions and show that complementary sources of information can boost the performance.
We provide a thorough ablation study on the benefits of different architectural as well as algorithmic design decisions.
Our method yields competitive results on the popular ScanNet benchmark and SceneNN dataset.
\end{abstract}

%% file: sec/01_introduction.tex
\section{Introduction}

\begin{figure}[t]
	\includegraphics[width=\linewidth]{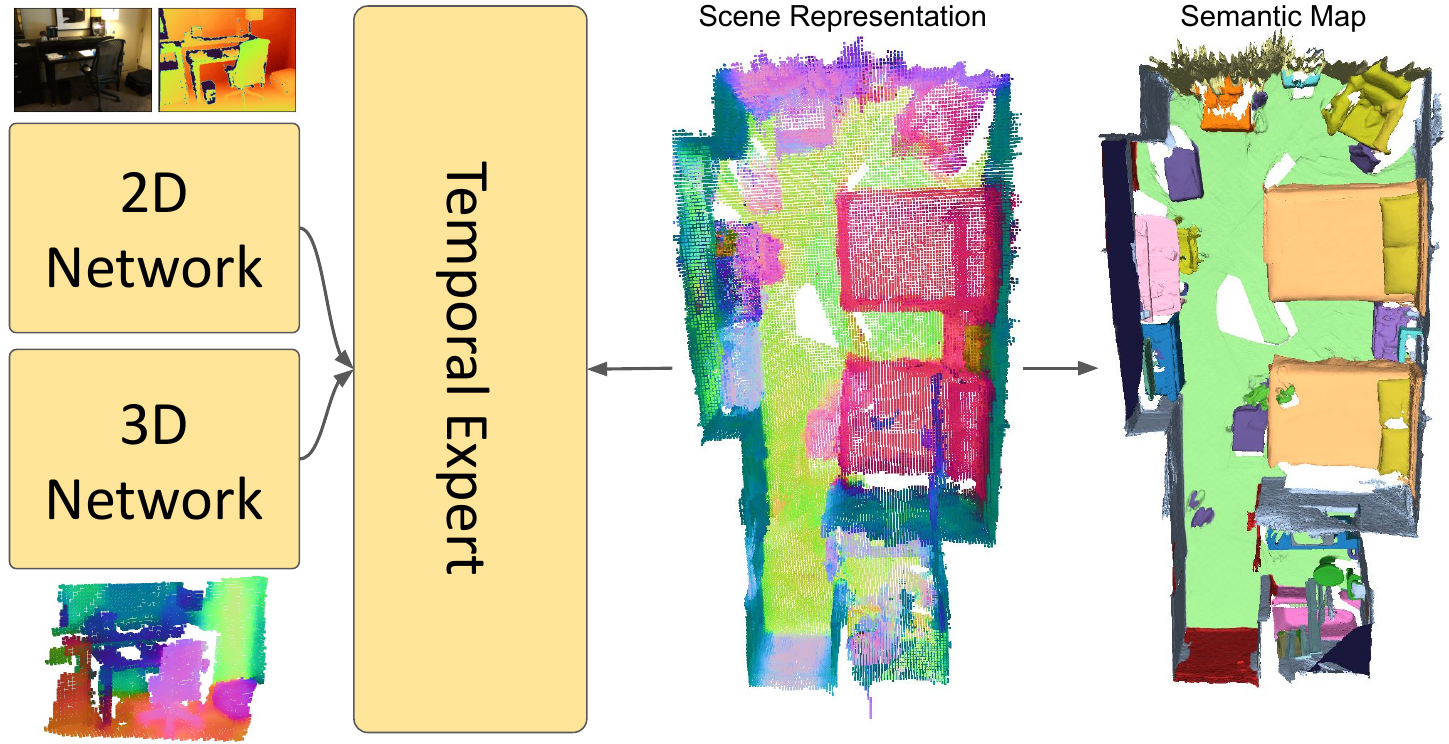}\\[-15pt]
	\caption{We propose an {online semantic 3D} reconstruction pipeline, which fuses RGB-D observations into a globally consistent semantic map. The key component is a local spatio-temporal expert network that fuses new observations into a learned scene representation. This temporal expert learns to select information from 2D, 3D, and previous steps using an attention mechanism. 
 }
\end{figure}


To interact with the world, humans not only require low-level spatial awareness of their surroundings, but also on real-time higher-level semantic understanding. 
Building such spatial awareness through 3D reconstruction has been a long-standing topic in computer vision.
In this work, we address the task of online 3D semantic reconstruction.
Specifically, given an incoming stream of RGB-D frames,
the goal is to reconstruct a semantically enriched 3D scene representation that is continuously updated.

The world around us is complex and 3D scenes can be understood at different levels -- 
from object-level understanding~\cite{mask3d}, over humans \cite{takmaz20233d} and full rooms \cite{yue2023connecting, chen2023polydiffuse} to large-scale outdoor spaces \cite{kreuzberg20224d},
both in carefully selected semantic classes \cite{Nekrasov-et-al-3DV-2021, Kundu-et-al-ECCV-2020} or more abstract concepts and affordances \cite{Peng2023OpenScene, takmaz2023openmask3d}.
While high-level understanding can be sufficient for planning and navigation,
detailed scene interactions require more fine-grained understanding with accurate semantic boundaries.
As such, fine-grained semantic understanding is at the heart of algorithms enabling real-world interactions.


Numerous works in 3D scene understanding focus on offline reconstructed point-clouds \cite{Qi-et-al-CVPR-2017, Kundu-et-al-ECCV-2020}, meshes \cite{Schult20CVPR}, or voxel-grids \cite{Nekrasov-et-al-3DV-2021, Choy-et-al-CVPR-2019}.
While there has been impressive progress with these works in recent years, a large majority of these have one major shortcoming that we aim to address. 
All these works require an a priori reconstruction of the scene and use this global information for the understanding task.
Therefore, they are considered offline methods.
However, autonomous agents (as well as humans) typically build a ``mental" map of the environment in an incremental manner, that is, continuously update it over time as new information is collected. 
Thus, scene understanding must inherently be an iterative process, as an autonomous agent cannot assume all information known a priori.
In this work, we investigate this particular problem, where we incrementally build a semantic map given a stream of posed RGB-D data that allows for online processing of the incoming data streams and can be integrated in real-time systems.
This online processing is essential for enabling real-world applications, such as robotics and mixed reality, where an updated semantic map is required to solve complex tasks.

Only few approaches tackle the problem specified above. 
The seminal works of Vineet~\etal\cite{Vineet-et-al-ICRA-2015} and SemanticFusion~\cite{Mccormac-et-al-ICRA-2017} map 2D semantic predictions into 3D.
One step further, PanopticFusion~\cite{narita2019panopticfusion} predicts a semantic instance map in 2D that is mapped and aggregated in 3D. 
While these methods reason in 2D as well as 3D, the 3D reasoning is a CRF-based regularizer that is limited compared to modern neural networks.
Further, the CRF requires global information of the entire scene that limits the scalability of the methods to small scenes. 
Similarly, INS-Conv~\cite{liu2022ins} estimates semantic instance maps using 3D processing with a large UNet that requires global processing to avoid drifting errors.
In contrast, other works \cite{zhang2020fusion,huang2021supervoxel} perform 3D reasoning using point- or supervoxel convolutions in a local frame.
However, they only store explicit labels that only encode per-point information, while our work uses learned features encoding low-, mid- and high-level information.

Our work is based on the observation that 2D and 3D information is complementary for the task of scene understanding. 
Some elements are better to be understood in 2D depending on context and geometry whereas others are easier to be segmented in 3D given their spatial structure. 
To this end, we present a novel attention-based aggregation mechanism that fuses 2D, 3D, as well as existing features into the scene.
Our method only operates in a local region defined by the new measurement and integrates the updates into the learned global scene representation.
Through this design, our method is independent of the scene size and can scale to large-scale scenes.

In an extensive experimental evaluation, we show that our method is competitive with existing approaches to online semantic 3D reconstruction while not requiring passes over the entire reconstruction as opposed to some other methods~\cite{liu2022ins}.
This is particularly important for online processing on mobile devices and agents that are constrained in the amount of compute and memory available.
We evaluate our method on ScanNet as well as SceneNN and present in-depth ablation studies to motivate our design choices.
We will release the source code on acceptance of this paper to foster further research in this direction.
In summary, the key contributions in this work are:
\begin{itemize}
    \item We show that 2D and 3D information are complementary for the task of online 3D semantic reconstruction and improve the overall result.
    \item We propose a novel local fusion approach that leverages an attention mechanism to combine existing features with new 2D and 3D information in an online fashion.
    We evaluate our pipeline design on the well-known ScanNet~\cite{Dai-et-al-CVPR-2017} benchmark and show competitive results compared to existing online local reconstruction methods.
\end{itemize}

%% file: sec/02_relatedwork.tex
\section{Related Work}

\boldparagraph{Offline \vs online processing.}
Most existing 3D segmentation methods follow on offline approach:
the 3D geometry of the scene and its corresponding features (color, normals, \etc{}) are known a-priori and then processed by the segmentation method.
We first review prior work on \emph{offline} semantic segmentation and then look at existing \emph{online} methods in the context of incrementally building semantic 3D maps.

\boldparagraph{3D Semantic Segmentation} is the problem of assigning a class label to each point, voxel, or vertex of the 3D scene.
It is central to many applications and pipelines that require some form of understanding.
In recent years, many different methods tackled this problem. 
Semantic Stixels~\cite{Schneider-et-al-IV-2016} predict 2D semantic labeling and stereo depth maps that are aggregated in a 3D stixel representation.
While this representation can be sufficient for outdoor applications, it lacks representation power for indoor applications. 
Kundu~\etal~\cite{Kundu-et-al-ECCV-2020} address the problem of lack of context in the 2D views by rendering views from an already reconstructed mesh to have a larger field-of-view that improves the performance of 2D semantic segmentation. 
The predictions are afterwards aggregated again on the 3D mesh.
As this approach is dependent on an already reconstructed mesh, it is not suitable for an online fusion approach.
Atlas~\cite{Murez-et-al-ECCV-2020} jointly reconstructs a semantic and geometric map from visual inputs by learning multi-view fusion. 
As this approach needs to aggregate dense viewing frustums to solve the multi-view stereo problem, it is not suitable for fast online updates.
SemanticNeRF~\cite{Zhi-et-al-ICCV-2021} proposes the application of recently proposed neural radiance fields~\cite{Mildenhall-et-al-ECCV-2020} to the problem of 3D semantic segmentation. 
While this approach shows impressive results, it is also not applicable to fast and accurate online updates of a semantic map.
Mix3D~\cite{Nekrasov-et-al-3DV-2021} boosts the performance of 3D segmentation methods by proposing a novel data augmentation technique that combines different scenes.
While this augmentation works for global methods, it cannot be applied to online fusion systems since we jointly learn the fusion across time and segmentation of the scene.
BPNet~\cite{Hu-et-al-CVPR-2021} couples 2D and 3D predictions of scene labels by proposing a bi-directional projection module. 
This boosts the performance on 3D semantic segmentation but is dependent on global processing and a-prior scene reconstructions.
VMNet~\cite{Hu-et-al-ICCV-2021} combines Euclidean and geodesic information to address the short-comings of voxel-only approaches. 
Yet, it also requires global processing to unfold its full potential.
OccuSeg~\cite{Han-et-al-CVPR-2020} enhances supervoxel-based geometric segmentation with learned features and refines them using graph-based clustering but requires a global receptive field, which makes them unnecessarily expensive for online processing.

\boldparagraph{Online 3D Semantic Segmentation.}
In contrast to the previously mentioned approaches, online methods iteratively process the scene making them better suitable to real-time applications, where agents are interacting with their environment such as robotics or mixed reality in unknown environments.
There is a long line of work aiming at the real-time reconstruction of geometry and appearance~\cite{Curless-et-al-SIGGRAPH-1996,Newcombe-et-al-ISMAR-2011,Weder-at-al-CVPR-2020,Weder-et-al-CVPR-2021}.
These works have been extended to scene understanding to enable agents with understanding capabilities.
The approaches \cite{Vineet-et-al-ICRA-2015} and \cite{Mccormac-et-al-ICRA-2017} proposed to fuse 2D semantic predictions into a global semantic map that is refined using a conditional random field (CRF).
This idea has been extended by several works.
SceneCode~\cite{Zhi-et-al-CVPR-2019} stores a per-keyframe latent code encoding the semantic information of the scene that is optimized at test time.
Meanwhile, MaskFusion~\cite{Runz-et-al-ISMAR-2018} and Fusion++~\cite{McCormac-et-al-3DV-2018} focus on 3D object segmentation while ignoring their semantic class.
ProgressiveFusion~\cite{pham2019progressivefusion} improves efficiency by clustering voxels into supervoxels and apply CRF on that level.
SemanticReconstruction~\cite{jeon2018semantic} follows a similar approach as~\cite{Mccormac-et-al-ICRA-2017}, but shows that their scene representation can be used for downstream tasks such as scene completion and manipulation.
PanopticFusion~\cite{narita2019panopticfusion} estimates 3D semantic instance maps by predicting 2D semantic and instance segmentation using off-the-shelf networks, aggregates them in 3D, and also regularizes them using a CRF.
While these works leverage 2D processing in combination with optimization-based 3D regularization, they all resort to traditional voxel fusion and do not utilize trainable 3D neural networks.
This shortcoming has been addressed in SVCNN~\cite{huang2021supervoxel}, which clusters voxels that store explicit semantic information into supervoxels, and then processes them using a special convolutional operator designed for supervoxels.
However, \cite{huang2021supervoxel} still resorts to an explicit fusion of 2D semantic information into voxels.
An alternative is FusionAware~\cite{zhang2020fusion} that represents scenes using efficient point cloud representations and uses point-convolutions to aggregate new information.
More recently, Liu~\etal~\cite{liu2022ins} presented an online method predicting semantic instance maps using only 3D processing.
Nevertheless, these two works disregard useful 2D information. 
In our work, we address these limitations by a) combining 2D and 3D information in a temporal expert network leveraging both sources of information, and b) applying a powerful yet lightweight 3D network on the current viewing frustum.

%% file: sec/03_method.tex
\section{Method}

This section presents our method for online 3D semantic reconstruction.
Firstly, we give an overview of our model (Fig.~\ref{fig:pipeline}) and the 3D scene representation.
Then, we describe 
the spatial-temporal expert that enables efficient local updates of the scene representation.
Lastly, we discuss training protocols, loss functions and sequential optimization.


\subsection{Semantic 3D Reconstruction Pipeline}

\begin{figure*}[t]
    \centering
	\includegraphics[width=\textwidth]{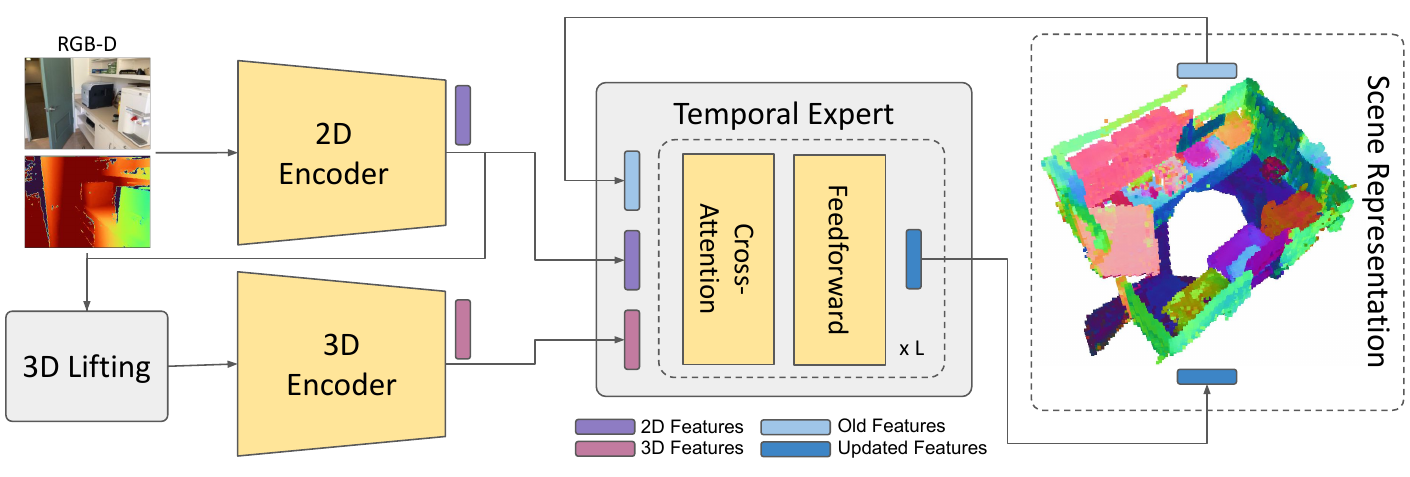}\\[-10pt]
	\caption{\textbf{Pipeline Overview:} Our pipeline consists of three main stages. The 2D encoder extracts information from incoming RGB-D imagery. This information is enhanced with 3D information using a light-weight 3D encoder. The information from these two sources is combined with existing information in the learned scene representation using the temporal expert.} \label{fig:pipeline}
\end{figure*}

\boldparagraph{Overview.}
Our proposed model consists of three major components (Fig.~\ref{fig:pipeline}) and a learned 3D scene representation. 
The first stage is a 2D encoder $F^\text{2D}$ that extracts 2D feature maps from an incoming stream of RGB-D images. 
The second stage is a 3D encoder $F^\text{3D}$ that incorporates 3D geometry into each feature map after lifting it to 3D using the given camera parameters and depth maps.
The third stage is a new \emph{temporal expert} network $F^{\Delta T}$ that consolidates 3D scene representations using complementary information from 2D and 3D as well as the so-far reconstructed 3D scene.
 
\boldparagraph{Scene Representation.}
The backbone of every online reconstruction method is a suitable scene representation. 
Typically, the primary choice are voxels, points, meshes or implicit (neural) representation. 
Meshes and implicit representations are difficult to update with new observations, while points lack information about geometric connectivity. 
This is crucial for scene understanding where decisions about segmentation boundaries are oftentimes guided by geometric boundaries.
Therefore, we represent scenes using a hybrid representation $S$ combining learned and explicit features that are stored in a sparse voxel grid.
Sparse voxel grids allow for efficient processing using neural networks. 
In particular, each voxel stores a learned feature $F$ of dimension $D_F$\,=\,$40$ encoding the aggregated information about the scene content.
This learned scene representation allows to store high-, mid-, and low-level information useful for the semantic segmentation task. 
This mitigates the need for expensive re-processing in deep neural networks at every time step to fuse existing and new information.
The voxels also store the number of per-voxel observations, which is relevant for the subsequent fusion step.

\boldparagraph{2D Encoder.} 
The aim of the first stage is to extract semantic features from incoming 2D RGB-D images using a 2D convolutional network $F^\text{2D}$.
The 2D network $F^\text{2D}$ with trainable parameters $\theta^\text{2D}$ takes RGB-D frames $(I_t, D_t)$ as input and predicts semantic features $\tilde{f}^\text{2D}_t$ per frame:
\begin{equation}
    \tilde{f}^\text{2D}_t = F^\text{2D}\left(\left[I_t, D_t, N_t\right]; \theta^\text{2D}\right)
\end{equation}
The normal map $N_t$ is estimated from the depth map $D_t$ and serves as additional input.
The network consists of DeepLabV3+~\cite{Chen-et-al-Arxiv-2017} and uses, similar to~\cite{Kundu-et-al-ECCV-2020}, an Xception65~\cite{Chollet-CVPR-2017} encoder that is adjusted to handle RGB-D-N input data (color, depth, normals) since geometric information improves semantic segmentation~\cite{Gupta-et-al-ECCV-2014, Wang-et-al-ECCV-2016, Hazirbas-et-al-ACCV-2016}.
The semantic features $\tilde{f}^\text{2D}_t$ are directly obtained from the DeepLab decoder. 
However, the original dimension of the features $\tilde{f}^\text{2D}_t$ is $D_{\tilde{f}^\text{2D}}$\,=\,$256$
which is too memory intensive for online processing of large indoor scenes.
Instead, we project the feature maps to ${D}^\text{2D}$\,=\,$40$.
This compression allows online processing for the remaining of the pipeline while still retaining all relevant information needed for 3D semantic understanding.
The 2D network  $F^\text{2D}$ is pre-trained on ImageNet~\cite{Deng-et-al-CVPR-2009} and fine-tuned on 2D training data from ScanNet~\cite{Dai-et-al-CVPR-2017}.
During the training of the full pipeline, the 2D encoder is partially fine-tuned using an auxiliary semantic segmentation head enforcing consistent performance across frames and for regularization of the joint feature space.

\boldparagraph{3D Encoder.}
We additionally process the incoming information in 3D as geometry is complementary to 2D appearance. 
This processing is particularly motivated by the possible reasoning about hidden geometric object boundaries occluded in the current 2D frame.
To this end, we lift the obtained 2D feature map ${f}^\text{2D}_t$ to 3D point clouds by projecting the depth map $D_t$ using the known, gravity-aligned camera orientation $R \in SO(3)$ and intrinsics. 
Due to noisy depth estimates, we additionally filter out points that are more than $3$\,m away from the camera.
The resulting \emph{local} 3D feature volume is refined using a light-weight U-Net~\cite{Ronneberger-et-al-MICCAI-2015} $F^\text{3D}$ yielding a 3D feature map $f^\text{3D}_t$ with the same feature dimension $D_F = 40$ as the 2D feature map $f^\text{2D}_t$.

\boldparagraph{Spatio-Temporal Expert.}
In the previous stages, 2D features are extracted and enhanced with 3D information.
In the next step, this information is integrated into the existing global scene representation $S_{t-1}$.
To this end, we propose the spatio-temporal expert network $F^{\Delta T}$ with weights $\theta^{\Delta T}$.
The task of the spatio-temporal expert network $F^{\Delta T}$ is to update the features stored in the learned scene representation given the new information from $f^\text{2D}_t$, $f^\text{3D}_t$, and the existing information $f^{\text{global}}_{t-1}$.
The feature volume $f^{\text{global}}_{t-1}$ is a crop of the relevant local sub-volume from the global scene representation $S_{t-1}$ using the known camera pose $\left[R | t\right] \in SE(3)$.
The overall mapping is computed as:
\begin{equation}
    f^{\text{global}}_{t} = F^{\Delta T}\left(\left[ f^{\text{global}}_{t-1}, f^\text{3D}_t, f^\text{2D}_t\right]; \theta^{\Delta T}\right)
\end{equation}
The resulting local volume $f^{\text{global}}_{t}$ is then written back, using the inverse camera pose, to obtain the new global scene representation $S_{t}$.
By providing access to both the 2D and 3D features in parallel, the expert network can learn where it is beneficial to rely more on 2D appearance features or where it is advantageous to trust the 3D geometry-based features, see Fig.~\ref{fig:attention} for an illustration. The 3D features reveal geometrical details while the 2D features provide textural information in flat areas with little geometric information.

\begin{figure}[t]
\centering
\includegraphics[width=\columnwidth]{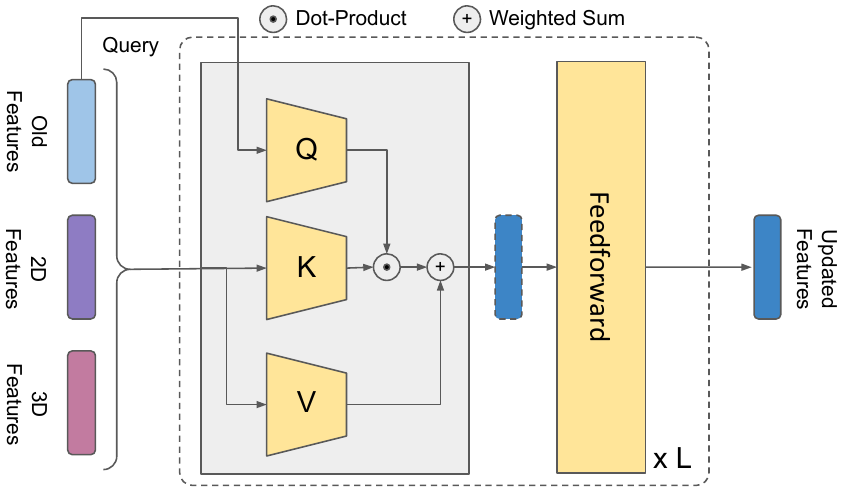}
\vspace{-20px}
\caption{\textbf{Temporal expert network.} The temporal expert network takes the three features ($f^{\text{global}}_{t-1}$, 2D, and 3D) as input, and iteratively refines the old feature vector to obtain the update feature that can be stored in the scene representation. The old feature is used as the query in the attention mechanism.}
\label{fig:fusion_network}
\end{figure}

$F^{\Delta T}$ is implemented as a Transformer consisting of cross-attention and feed-forward layers (see Fig.\,\ref{fig:fusion_network}).
The task of the cross-attention layer is to extract relevant information from the three sources of information ($f^\text{2D}_t$, $f^\text{3D}_t$, $f^{\text{global}}_{t-1}$) using $f^{\text{global}}_{t-1}$ as query features. 
The attention is defined as
$
    f = w_{f^{\text{global}}_{t-1}} v_{f^{\text{global}}_{t-1}} + w_{f^\text{2D}_t} v_{f^\text{2D}_t} + w_{ f^\text{3D}_t} v_{ f^\text{3D}_t}
$
%
and the weights $(w_{f^{\text{global}}_{t-1}}, w_{f^\text{2D}_t},w_{ f^\text{3D}_t})$ are defined as: 
\begin{equation}
   w_{f^k} = Q\left(f^{\text{global}}_{t-1}\right)^T K\left(f^{k}\right) \enspace,
\end{equation}
where $f^k \in \{f^{\text{global}}_{t-1}, f^\text{3D}_t, f^\text{2D}_t \}$.
The values $v_{f^k}$ are obtained using a linear projection layer $v_{f^k} = V\left(f^{k}\right)$.

The features extracted from the cross-attention layer are first normalized using layer norm and then refined using a standard feed-forward layer. 
Furthermore, both layers are augmented with a skip connection guaranteeing healthy gradients during training.
The refined features that consist of information extracted from the three sources $f^\text{2D}_t$, $f^\text{3D}_t$, and $f^{\text{global}}_{t-1}$ are written back into the global scene representation. 
The temporal expert is also supervised by a point-wise segmentation loss,
ensuring optimal segmentation given the currently available 3D scene information.

\input{figures/vis_attention}

\boldparagraph{Loss Function.}
The pipeline is trained using focal loss~\cite{focalloss} at several stages in the pipeline. 
These losses are applied after the 2D encoder $F^\text{2D}$, the 3D encoder $F^\text{3D}$, and the temporal expert network $F^{\Delta T}$.
These auxiliary supervision signals are required to constrain the feature space that encodes the information throughout the entire pipeline.
Further, these auxiliary losses ensure that each stage solves the task of semantic segmentation as good as possible for themselves providing the temporal expert network with valuable information.
%
%
The overall loss is the sum of these losses:
\begin{equation}
    \mathcal{L} = \lambda_{\text{2D}} \mathcal{L}_{\text{2D}} + \lambda_{\text{3D}} \mathcal{L}_{\text{3D}} + \lambda_{\text{Expert}} \mathcal{L}_{\text{Expert}}
\end{equation}
where each term $\mathcal{L}_{\text{2D}}, \mathcal{L}_{\text{3D}}, \mathcal{L}_{\text{Expert}}$ is a focal loss, defined as:

\begin{equation}
    \mathcal{L} = (1 - \hat{y})^{\gamma}\text{CE}(\hat{y}, y)
\end{equation}
As the loss is applied on a per-voxel level, the corresponding ground-truth labels first need to be mapped from the ground truth polygon mesh to a voxelized representation.
To this end, we first voxelize all scenes using the target resolution and assign the label of the closest vertex of the mesh.
Closest points are efficiently found using KD-tree-based nearest neighbor search.

\boldparagraph{Sequential Training.}
The temporal expert network needs to learn how to fuse new information into the existing scene representation based on sequential data. 
A key challenge is catastrophic forgetting, where the network forgets what it has learned during the beginning of a sequence and only focuses on the last few frames along a camera trajectory.
To overcome this challenge, it is critically important to randomly select camera views along each video trajectory, \ie{}, a permutation of the original frame order.
Similarly, to avoid that the model only sees fully reconstructed scenes after some initial training time, we randomly reset the reconstructed scenes so that the model always sees scenes at varying levels of reconstruction.

%% file: figures/vis_attention.tex
\begin{figure}[tb]
\centering
\footnotesize
\setlength{\tabcolsep}{0.5px}%
\begin{tabular}{cccc}
    \textbf{2D Labels} & \textbf{2D Attention} & \textbf{3D Label} & \textbf{3D Attention} \\
    \begin{tikzpicture}
    \node[anchor=south west,inner sep=0] (image) at (0,0) {  \includegraphics[width=0.24\columnwidth, height=0.24\columnwidth]{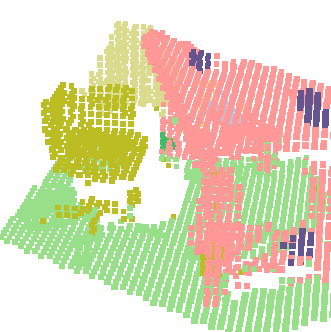}};
    \draw[red,ultra thick,rounded corners] (0.,0.3) rectangle (1.,1.);
    \end{tikzpicture}
    & 
    \begin{tikzpicture}
    \node[anchor=south west,inner sep=0] (image) at (0,0) {  \includegraphics[width=0.24\columnwidth, height=0.24\columnwidth]{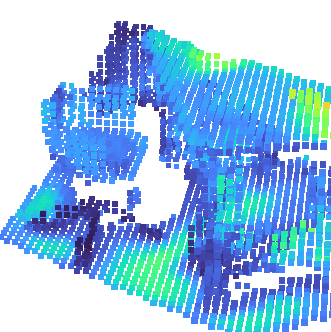}};
    \draw[red,ultra thick,rounded corners] (0.,0.3) rectangle (1.,1.);
    \end{tikzpicture}
    & 
    \begin{tikzpicture}
    \node[anchor=south west,inner sep=0] (image) at (0,0) {  \includegraphics[width=0.24\columnwidth, height=0.24\columnwidth]{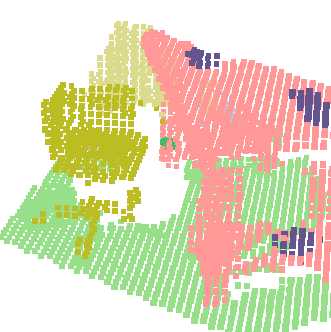}};
    \draw[red,ultra thick,rounded corners] (0.,0.3) rectangle (1.,1.);
    \end{tikzpicture}
    & 
   \begin{tikzpicture}
    \node[anchor=south west,inner sep=0] (image) at (0,0) {  \includegraphics[width=0.24\columnwidth, height=0.24\columnwidth]{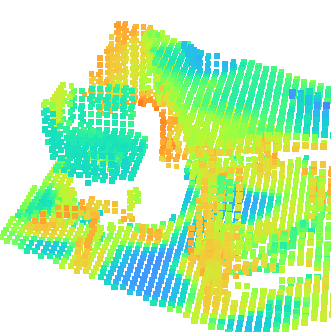}};
    \draw[red,ultra thick,rounded corners] (0.,0.3) rectangle (1.,1.);
    \end{tikzpicture}
    \\
    \begin{tikzpicture}
    \node[anchor=south west,inner sep=0] (image) at (0,0) {  \includegraphics[width=0.24\columnwidth, height=0.24\columnwidth]{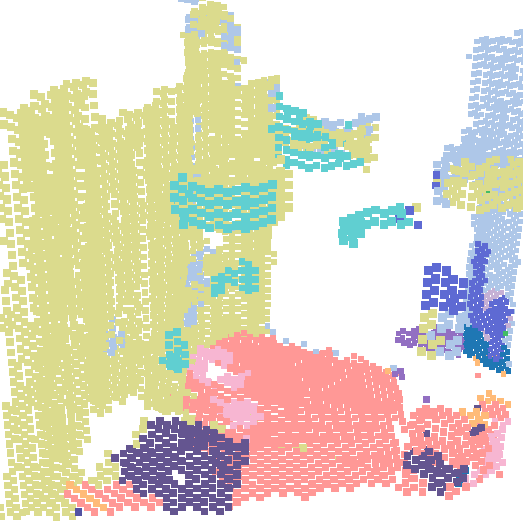}};
    \draw[red,ultra thick,rounded corners] (.5,1.) rectangle (1.6,1.7);
    \end{tikzpicture}
    &  
    \begin{tikzpicture}
    \node[anchor=south west,inner sep=0] (image) at (0,0) {  \includegraphics[width=0.24\columnwidth, height=0.24\columnwidth]{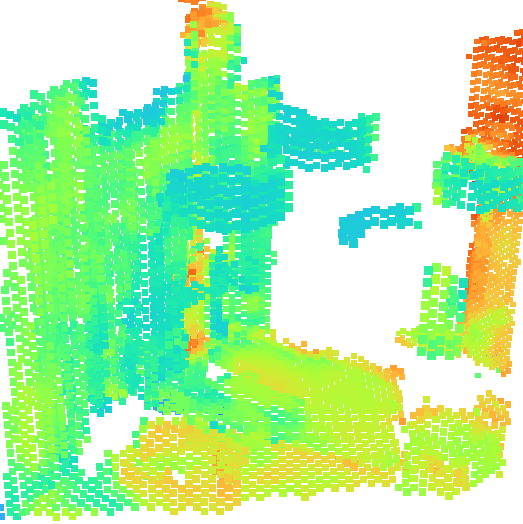}};
    \draw[red,ultra thick,rounded corners] (.5,1.) rectangle (1.6,1.7);
    \end{tikzpicture}
    & 
    \begin{tikzpicture}
    \node[anchor=south west,inner sep=0] (image) at (0,0) {  \includegraphics[width=0.24\columnwidth, height=0.24\columnwidth]{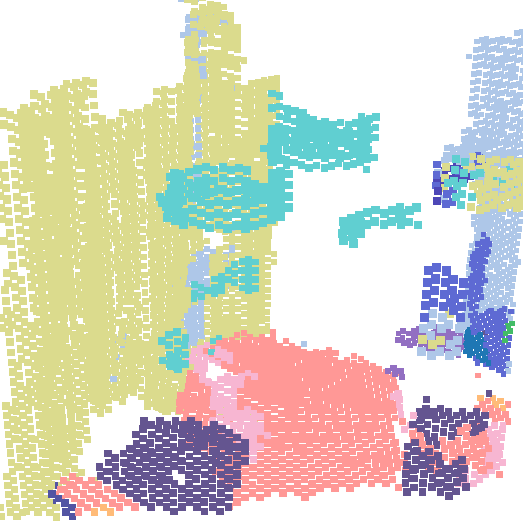}};
    \draw[red,ultra thick,rounded corners] (.5,1.) rectangle (1.6,1.7);
    \end{tikzpicture}
    &
     \begin{tikzpicture}
    \node[anchor=south west,inner sep=0] (image) at (0,0) {  \includegraphics[width=0.24\columnwidth, height=0.24\columnwidth]{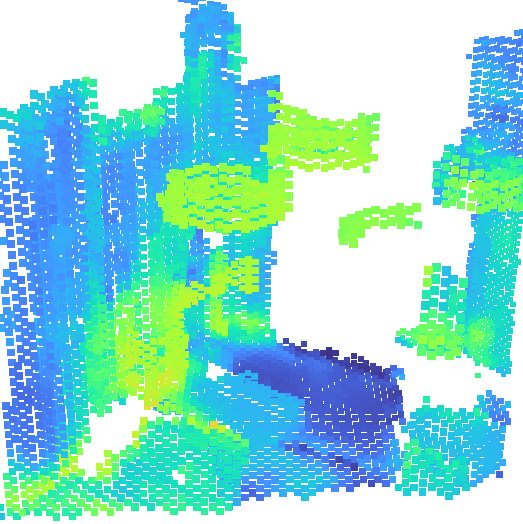}};
    \draw[red,ultra thick,rounded corners] (.5,1.) rectangle (1.6,1.7);
    \end{tikzpicture}
    \\
\end{tabular}
\vspace{-6px}
\caption{\textbf{Temporal expert attention maps}. We visualize the attention maps for the 2D and 3D input features in the expert network. The expert attends to the 3D features to refine the segmentation of fine details (legs of the chair, lamps), while it attends to the 2D features to predict the label for large areas (table, walls, \etc). \vspace{-10px}
}
\label{fig:attention}
\end{figure}

%% file: sec/04_experiments.tex
\section{Experiments}

\input{tables/scannet_benchmark}


\subsection{Implementation Details}
We implement the proposed pipeline in PyTorch.
We use the MinkowskiEngine\,\cite{Choy-et-al-CVPR-2019} for the sparse 3D convolutions in the 3D encoder, and Pytorch3D~\cite{ravi2020pytorch3d} for the geometric projections.
The entire pipeline is trained with the Adam optimizer and a OneCycle~\cite{smith2017onecycle} learning rate scheduler.
Due to memory constraints, the batch size is $4$ but we obtain an effective batch size of 8 by aggregating the gradients across two batches.
We set the maximum learning rate to $0.001$ for the 3D and temporal expert networks, while the maximum learning rate for the pre-trained 2D encoder is set to $1e-05$.
We equally weight the different terms in the loss function setting 
$\lambda_{\text{3D}}$\,=\,$\lambda_{\text{2D}}$\,=\,$\lambda_{\text{Expert.}}$\,=\,$1$. 
Further, we set the parameter of the focal loss $\gamma$\,=\,$1.$
We use five layers in the temporal expert transformer with a hidden dimension of $D_{\text{hidden}}$\,=\,$128$ in the feed-forward layers. 
The voxel grid resolution for the entire pipeline is set to $4$\,cm.

\input{figures/quali}

\subsection{Online Methods in Comparison}

\boldparagraph{FusionAware~\cite{zhang2020fusion}.}
Unlike our voxel-based representation, FusionAware is a \emph{point}-based online 3D semantic segmentation method.
The method aggregates measurements in 3D space using point convolutions and computes intra- and inter-frame features.

\boldparagraph{SVCNN~\cite{huang2021supervoxel}.}
Similar to ours, Supervoxel Convolution (SVCNN) is another candidate from the space of voxel-based approaches. 
SVCNN uses dedicated convolutional operators that operate directly on supervoxels and aggregate multi-view features during online reconstruction. 


\subsection{Datasets and Metrics}

\boldparagraph{ScanNet~\cite{Dai-et-al-CVPR-2017}}
consists of 1513 scans from 707 unique indoor scenes containing 2.5M RGB-D frames. 
All scenes provide dense 3D semantic annotations mapped to the NYU40 class labels.
Each scene is recorded up to three times using an iPad equipped with an Occipital depth sensor.
The camera poses and dense reconstruction of the scenes are obtained using BundleFusion~\cite{bundlefusion}.
The 3D labels are projected into all 2D frames to provide the 2D labels.

\boldparagraph{SceneNN~\cite{hua2016scenenn}.}
SceneNN consists of 76 scenes with semantic annotations and corresponding posed RGB-D data. 
We follow \cite{huang2021supervoxel, liu2022ins} and demonstrate the generalization capabilities of our method by training on ScanNet and evaluating on the 76 SceneNN scenes.

\input{tables/per_class_expert_2d_3d}
\input{tables/semantic_scannet_scenenn.tex}

\boldparagraph{Metrics.}
We follow the standard metrics of the ScanNet and SceneNN datasets.
In particular, we compute the mean and per-class intersection over union (IoU) on ScanNet, as well as the mean accuracy (mAcc) and weighted intersection over union (wIoU) on SceneNN.

\subsection{3D Semantic Segmentation}

Table~\ref{tab:scannet-benchmark} reports 3D semantic segmentation scores of our and recent methods on ScanNet~\cite{Dai-et-al-CVPR-2017}.
We compare online and offline methods, as well as local and global methods.
While offline methods rely on a pre-computed 3D scene reconstruction in the form of a point cloud or polygon mesh, online methods are able to reconstruct the 3D scene on the fly as new frames become available.
This functionality is attractive for online applications in robotics or AR/VR devices,
however they cannot rely on the full scene context which makes semantic reasoning harder and semantic scores are generally higher for offline methods \cite{Nekrasov-et-al-3DV-2021, Kundu-et-al-ECCV-2020, Choy-et-al-CVPR-2019}.
In the group of online methods, our approach improves over the existing local methods like SVCNN~\cite{huang2021supervoxel} and FusionAware~\cite{zhang2020fusion} by at least +$3.3$ mIoU.
Local methods operate on a local window defined by the newly incoming frames and are therefore memory and computationally efficient, both attractive qualities for real-time processing.
Global methods require global passes over the reconstructed scenes either by CRF regularization or neural network processing.
This step takes increasingly more time as the reconstructed scene becomes larger in size.
These methods are therefore less applicable for real-time applications, since
no upper bound on the processing time can be guaranteed.

In Table~\ref{tab:semantic_scannet_scenenn}, we compare our method to existing baselines on the ScanNet validation set as well as SceneNN. 
For ScanNet, we report the mIoU over all benchmark classes.
For SceneNN, we compute the weighted intersection over union (wIoU) and the mean accuracy (mAcc) for all annotated NYU40 classes.
In addition to the quantitative metrics, we also report the voxel resolution for all methods where it is available. A smaller voxel size generally results in better scores since finer details can be represented, however this comes at increased memory costs.
Our proposed method performs best among all local methods on both ScanNet and SceneNN (with a close second on the wIoU metric).
When also compared to global methods, INS-Conv~\cite{liu2022ins} performs only marginally better on ScanNet, even when using a voxel resolution that is twice as high, highlighting the memory efficiency of our method. 


\subsection{Ablation Studies}

\boldparagraph{Does the temporal expert network improve upon the individual 2D and 3D networks?}
In Table~\ref{tab:ablation_2d_3d_expert}, we report the numbers for the different stages in our pipeline. 
For each stage, we compute the per-class intersection over union on the ScanNet validation set.
The per-stage labels are obtained from the auxiliary heads used during training to constrain the joint feature space and aggregated using a simple voting mechanism. 
These labels are then mapped to the ScanNet ground-truth and evaluated using their evaluation pipeline.
The numbers show that the expert is consistently better than the individual branches (Ours - 2D and Ours - 3D). 
Further, the fact that sometimes the 2D labeling is better than the 3D labels and the significant margin between 3D and expert indicate that 1) bypassing the 2D information around the 3D encoder is useful and 2) our attention-based fusion mechanism allows better reasoning over time than simple voting.
We also show the differences between the different stages in Figure~\ref{fig:quali}, where one can see the benefits of selecting information from the two different encoders.

\boldparagraph{What does the temporal expert network attend to?}
In Figure~\ref{fig:attention}, we visualize the attention maps for different frames during the fusion process together with the corresponding predicted labels. 
We qualitatively show that they learn to leverage the two different encoder according to their individual strengths.
The temporal expert network attends to the 3D network for fine-details usually refining edges and geometric details (\eg, legs of a chair, edge of a table) while it attends to the 2D feature for information about large regions.
Further, we observe that the fusion with the old scene representation happens in later layers while earlier layers combine the 2D and 3D information.

\boldparagraph{What is the impact of the feature dimension $D_F$?}
The dimension $D_F$ of the features stored in the learned scene representation is a key hyperparameter of the pipeline.
Thus, we evaluate its impact on the overall performance in Table~\ref{tab:ablation_2d_3d_expert}. One can see that for $D_F$ below 40 (the default value), the performance is slightly deteriorated due to the required compression of the semantic information. For $D_F = 64$ the main reason for the slight performance drop is the increased overfitting to the training data due to the increased capacity of the features.  

\begin{figure}[ht]
\centering
\vspace{-2px}
\includegraphics[width=0.47\textwidth]{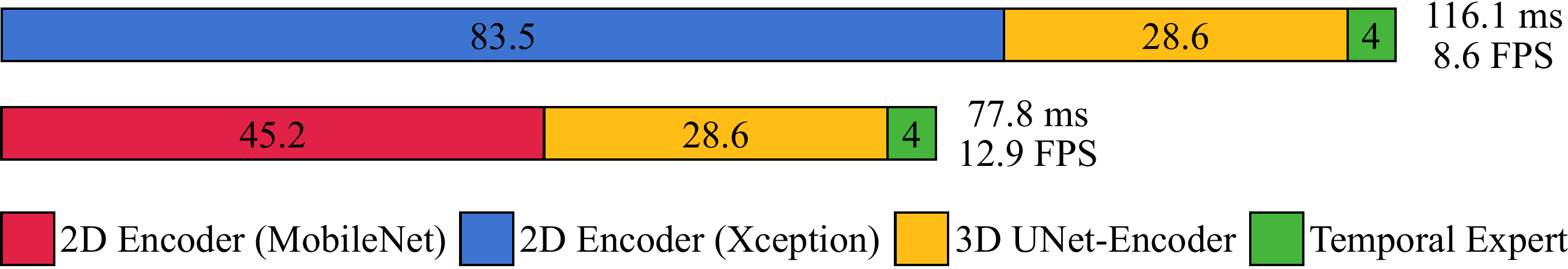}
\vspace{-5px}
\caption{\textbf{Runtime analysis for different 2D backbones.} A smaller 2D encoder trades runtime for accuracy (\cf~Tab.~\ref{tab:ablation_2d_3d_expert}). \vspace{-15px}}
\label{fig:runtime}
\end{figure}

\boldparagraph{How fast is our method?}
An average step through our entire (non-optimized) pipeline takes $116.1$\,ms ($8.6$ FPS) on an NVIDIA RTX\,2080 and $3.6$\,GHz Intel CPU i9-9900K.
To identify the main bottleneck, we analyse the runtime of the individual components in Fig.\,\ref{fig:runtime}.
One pass through the 2D network DeepLabV3~\cite{Chen-et-al-Arxiv-2017} plus lifting takes on average $83.5$\,ms. One pass through our light-weight 3D UNet takes $28.6$\,ms on average, and the temporal expert network operates at $4$\,ms per frame.
These numbers reveal the 2D DeepLabV3 as the main bottleneck. 
Thus, we also report the performance of MobileNet~\cite{mobilenet} in Tab.~\ref{tab:ablation_2d_3d_expert} instead of the standard Xception encoder.
This reduces the 2D processing time to $45.2$\,ms and boosts the overall runtime to $12.9$ FPS.

\boldparagraph{How many parameters does our pipeline have?}
Our pipeline consists in $51 \cdot 10^6$ parameters in total. 
The largest share is due to the the DeepLabV3 (Xception) model, which consists of $41 \cdot 10^6$  parameters.
The 3D U-Net consists of $10 \cdot 10^6$  parameters.
Compared to both encoders, the expert model takes a relatively small share of $92 \cdot 10^3$ parameters.
This further justifies our architecture design, with a marginal increase in model size and runtime, we obtain a notable boost in performance (+$1.6$ mIoU, see Tab.~\ref{tab:ablation_2d_3d_expert})


%% file: tables/scannet_benchmark.tex
\begin{table*}[ht!]
\setlength{\tabcolsep}{3px}%
\centering
\resizebox{\textwidth}{!}{
\begin{tabular}{c l c c c c c c c c c c c c c c c c c c c c c c}
& \textbf{Method}	&
\textbf{Processing} &
\textbf{mIoU}$\uparrow$	&
\rotatebox{70}{\textbf{Bathtub}}	&
\rotatebox{70}{\textbf{Bed}}	&
\rotatebox{70}{\textbf{Bookshelf}}
&	\rotatebox{70}{\textbf{Cabinet}}
&	\rotatebox{70}{\textbf{Chair}}
&	\rotatebox{70}{\textbf{Counter}}
&	\rotatebox{70}{\textbf{Curtain}}
&	\rotatebox{70}{\textbf{Desk}}
&	\rotatebox{70}{\textbf{Door}}
&	\rotatebox{70}{\textbf{Floor}}
&	\rotatebox{70}{\parbox{1cm}{\textbf{Other \\ Furniture}}}
&	\rotatebox{70}{\textbf{Picture}}
&	\rotatebox{70}{\textbf{Fridge}}
&	\rotatebox{70}{\parbox{1cm}{\textbf{Shower \\ Curtain}}}
&	\rotatebox{70}{\textbf{Sink}}
&	\rotatebox{70}{\textbf{Sofa}}
&	\rotatebox{70}{\textbf{Table}}
&	\rotatebox{70}{\textbf{Toilet}}
&	\rotatebox{70}{\textbf{Wall}}
&	\rotatebox{70}{\textbf{Window}}	\\
\midrule
\multirow{3}{*}{\rotatebox[origin=c]{90}{\textbf{Offline}}} & Mix3D~\cite{Nekrasov-et-al-3DV-2021}	& \textcolor{orange}{Global} &
    \textbf{78.1}	&	\textbf{96.4}	&	\textbf{85.5}	&	84.3	&	\textbf{78.1}	&	85.8	&	\textbf{57.5}	&	83.1	&	68.5	&	\textbf{71.4}	&	\textbf{97.9}	&	\textbf{59.4}	&	31.0	&	\textbf{80.1}	&	89.2	&	\textbf{84.1}	&	\textbf{81.9}	&	\textbf{72.3}	&	\textbf{94.0}	&	\textbf{88.7}	&	72.5	\\
& VirtualMVFusion~\cite{Kundu-et-al-ECCV-2020}	& \textcolor{orange}{Global} &	74.6	&	77.1	&	81.0	&	\textbf{84.8}	&	70.2	&	\textbf{86.5}	&	39.7	&	\textbf{89.9}	&	\textbf{69.9}	&	66.4	&	94.8	&	58.8	&	\textbf{33.0}	&	74.6	&	85.1	&	76.4	&	79.6	&	70.4	&	93.5	&	86.6	&	\textbf{72.8}	\\
& Minkowski~\cite{Choy-et-al-CVPR-2019} & \textcolor{orange}{Global}	&	73.6	&	85.9	&	81.8	&	83.2	&	70.9	&	84.0	&	52.1	&	85.3	&	66.0	&	64.3	&	95.1	&	54.4	&	28.6	&	73.1	&	\textbf{89.3}	&	67.5	&	77.2	&	68.3	&	87.4	&	85.2	&	72.7	\\
   \midrule
\multirow{5}{*}{\rotatebox[origin=c]{90}{\textbf{Online}}} 
& PanopticFusion~\cite{narita2019panopticfusion} & \textcolor{orange}{Global}
& 52.9 &	49.1 &	68.8 &	60.4 &	38.6 &	63.2 &	22.5 &	70.5 &	43.4 &	29.3 &	81.5 &	34.8 & 	24.1 & 	49.9	& 	66.9 &	50.7 &	64.9 &	44.2 &	79.6 &	60.2 &	56.1 \\ 
& INS-Conv~\cite{liu2022ins}	& \textcolor{orange}{Global}
& 71.7 &	75.1	&	75.9	&	81.2	&	70.4	&	86.8	&	53.7	&	84.2	&	60.9	&	60.8	&	95.3	&	53.4	&	29.3	&	61.6	&	86.4	&	71.9	&	79.3	&	64.0	&	93.3	&	84.5	&	66.3	\\
\cmidrule(lr){2-24}
& FusionAware~\cite{zhang2020fusion}	& \textcolor{Green}{Local} &	63.0	&	60.4	&	74.1	&	\textbf{76.6} 	&	59.0	&	74.7	&	50.1	&	73.4	&	50.3	&	52.7	&	91.9	&	45.4	&	32.3 	&	55.0	&	42.0	&	67.8	&	68.8	&	54.4	&	89.6	&	79.5	&	62.7	\\
& SVCNN~\cite{huang2021supervoxel} & \textcolor{Green}{Local}	&	63.5	&	65.6	&	71.1	&	71.9	&	61.3	&	75.7	&	44.4	&	\textbf{76.5}	&	53.4	&	56.6	&	92.8	&	47.8	&	27.2	&	\textbf{63.6}	&	53.1	&	66.4	&	64.5	&	50.8	&	86.4	&	79.2	&	61.1	\\
& \name{} (Ours)	& \textcolor{Green}{Local} &	\textbf{66.8}	&	\textbf{82.2}	&	\textbf{77.1}	&	49.6	&	\textbf{65.1}	&	\textbf{83.3}	&	\textbf{54.1}	&	76.1	&	\textbf{55.5}	&	\textbf{61.1}	&	\textbf{96.6}	&	\textbf{48.9}	&	\textbf{37.0}	&	38.8	&	\textbf{58.0}	&	\textbf{77.6}	&	\textbf{75.1}	&	\textbf{57.0}	&	\textbf{95.6}	&	\textbf{81.7}	&	\textbf{64.6}	\\
\bottomrule
\end{tabular}
}
\caption{
\textbf{3D Semantic Segmentation on ScanNet~\cite{Dai-et-al-CVPR-2017} Test.}
Offline baselines predict semantic labels using a-priori 3D scene reconstructions and \textcolor{orange}{global} passes over the entire scene. 
Online but \textcolor{orange}{global} baselines do online reasoning but require global passes over the full scene. 
\textcolor{Green}{Local} methods are online and reason only on local information within the viewing frustum and  on currently updated points.  
Among local methods, our proposed approach improves over existing baselines by at least +$3.3$ mIoU.}

\label{tab:scannet-benchmark}
\end{table*}

%% file: figures/quali.tex
\begin{figure*}[ht]
\centering
\footnotesize
\setlength{\tabcolsep}{1px}%
\newcommand{\sz}{0.18}
\vspace{15px}
\begin{tabular}{ccccc}
     \rotatebox{90}{\textbf{{~~~~~~~~~~~2D Labels}}}
    \begin{tikzpicture}
    \node[anchor=south west,inner sep=0] (image) at (0,0) { \includegraphics[height=\sz\textwidth]{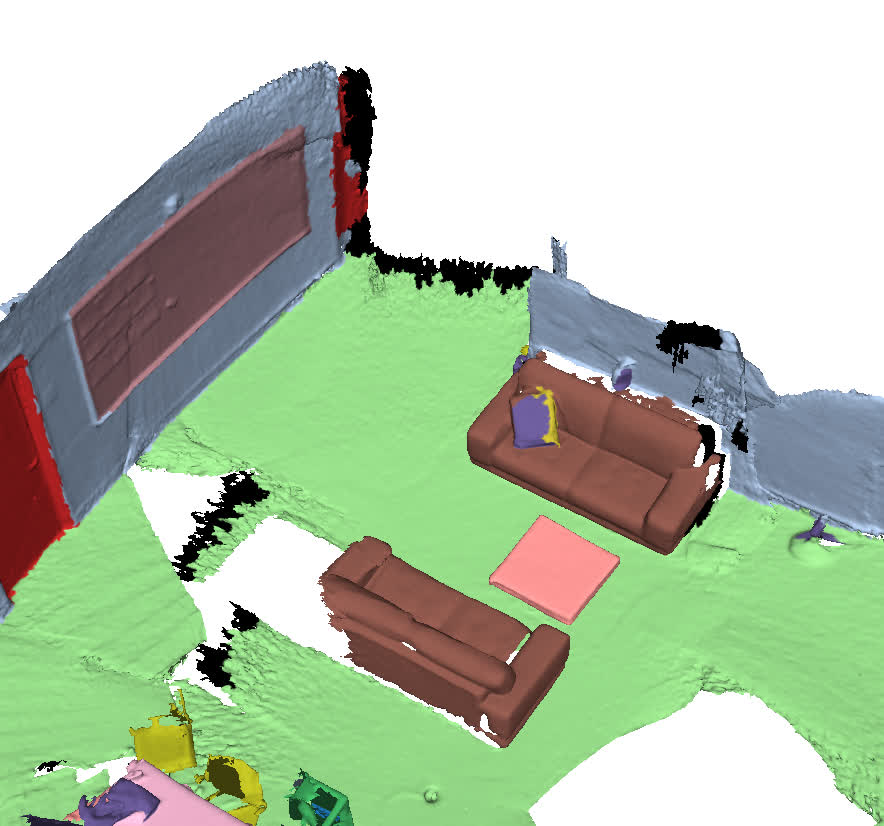}};
    \draw[red,ultra thick,rounded corners] (1.5,0.6) rectangle (2.8,2.);
    \end{tikzpicture}
    & 
    \begin{tikzpicture}
    \node[anchor=south west,inner sep=0] (image) at (0,0) { \includegraphics[height=\sz\textwidth]{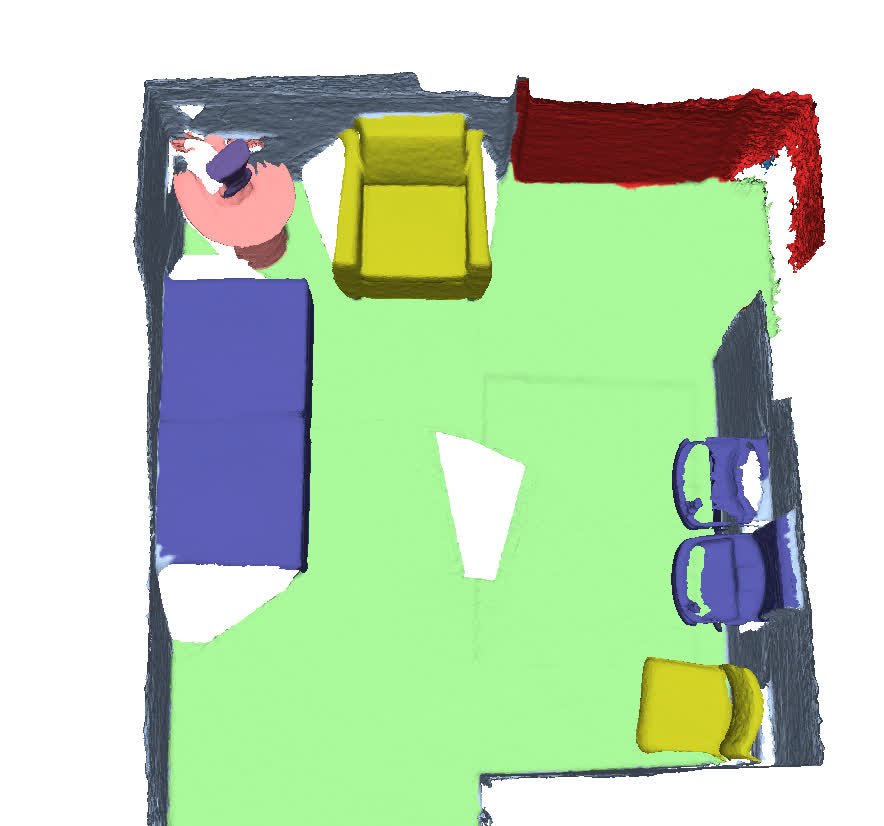}};
    \draw[red,ultra thick,rounded corners] (0.5,1.) rectangle (2.,3.);
    \end{tikzpicture}
    & 
     \begin{tikzpicture}
    \node[anchor=south west,inner sep=0] (image) at (0,0) { \includegraphics[height=\sz\textwidth]{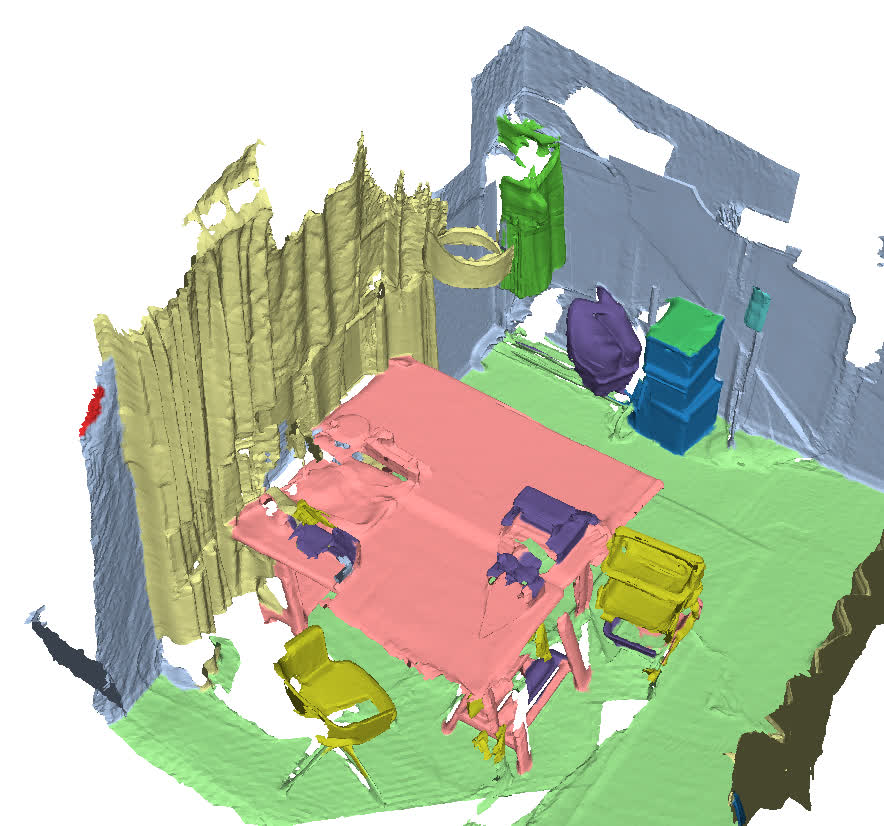}};
    \draw[red,ultra thick,rounded corners] (1.4,1.7) rectangle (2.4,2.9);
    \end{tikzpicture}
    & 
    \begin{tikzpicture}
    \node[anchor=south west,inner sep=0] (image) at (0,0) { \includegraphics[height=\sz\textwidth]{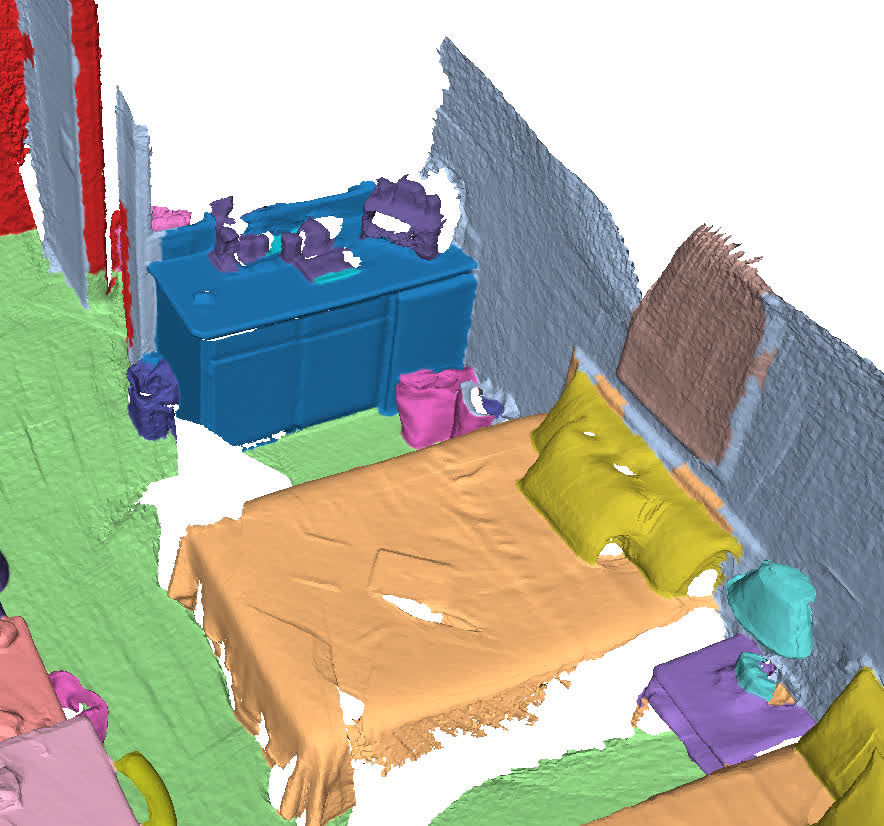}};
    \draw[red,ultra thick,rounded corners] (1.2,1.2) rectangle (2.3,2.5);
    \draw[red,ultra thick,rounded corners] (2.4,.1) rectangle (3.3,1.2);
    \end{tikzpicture}
    & 
    \begin{tikzpicture}
    \node[anchor=south west,inner sep=0] (image) at (0,0) { \includegraphics[height=\sz\textwidth]{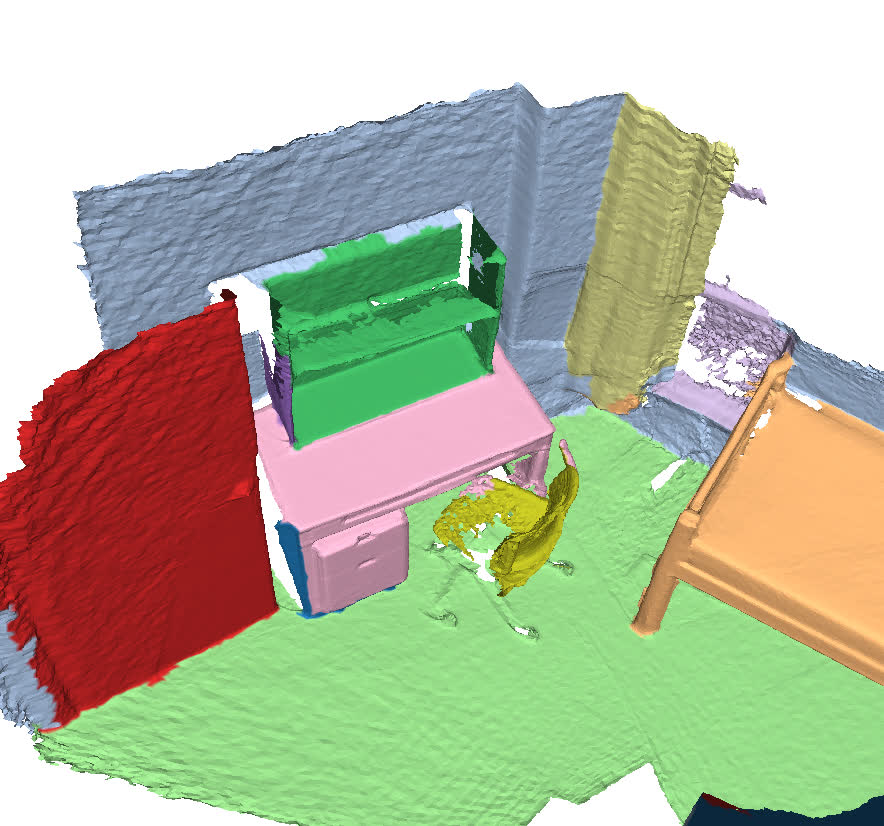}};
    \draw[red,ultra thick,rounded corners] (0.9,.6) rectangle (1.9,1.7);
    \end{tikzpicture}
    \\
    \rotatebox{90}{\textbf{{~~~~~~~~~~~3D Labels}}}    
    \begin{tikzpicture}
    \node[anchor=south west,inner sep=0] (image) at (0,0) { \includegraphics[height=\sz\textwidth]{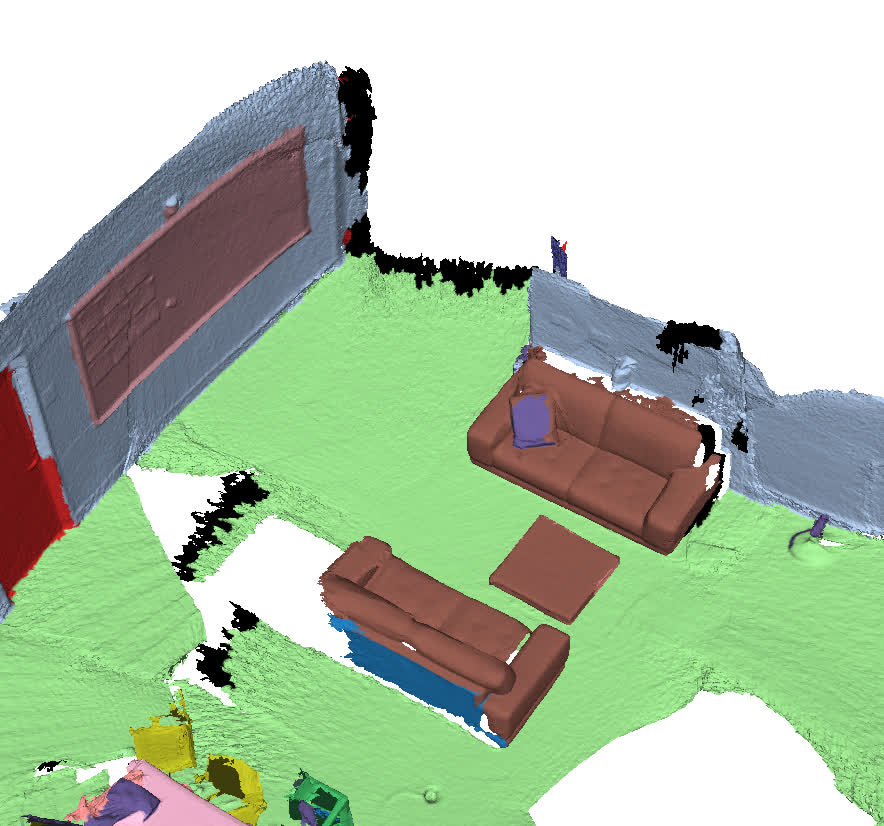}};
    \draw[red,ultra thick,rounded corners] (1.5,0.6) rectangle (2.8,2.);
    \end{tikzpicture} 
    &
    \begin{tikzpicture}
    \node[anchor=south west,inner sep=0] (image) at (0,0) { \includegraphics[height=\sz\textwidth]{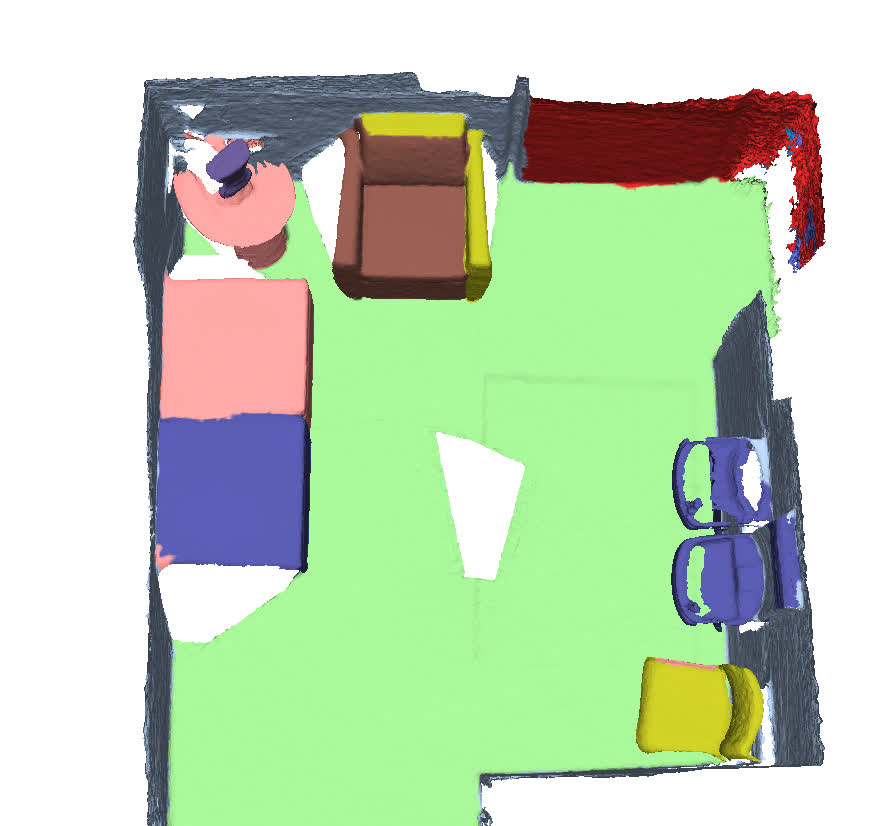}};
    \draw[red,ultra thick,rounded corners] (0.5,1.) rectangle (2.,3.);
    \end{tikzpicture}
    & 
    \begin{tikzpicture}
    \node[anchor=south west,inner sep=0] (image) at (0,0) { \includegraphics[height=\sz\textwidth]{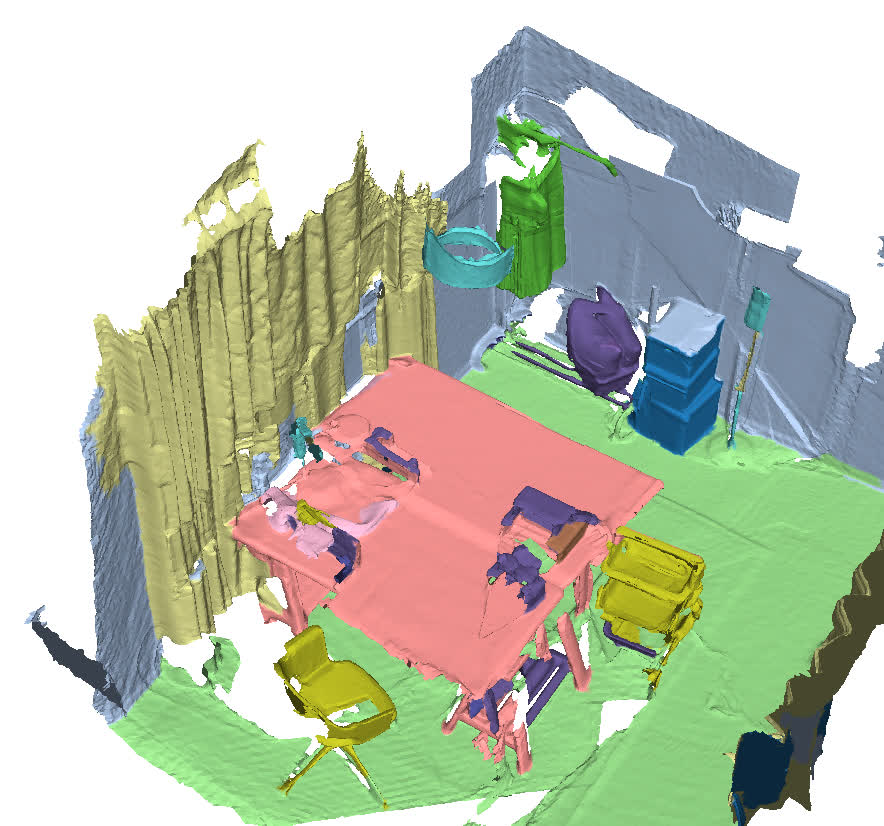}};
    \draw[red,ultra thick,rounded corners] (1.4,1.7) rectangle (2.4,2.9);
    \end{tikzpicture}
    & 
    \begin{tikzpicture}
    \node[anchor=south west,inner sep=0] (image) at (0,0) { \includegraphics[height=\sz\textwidth]{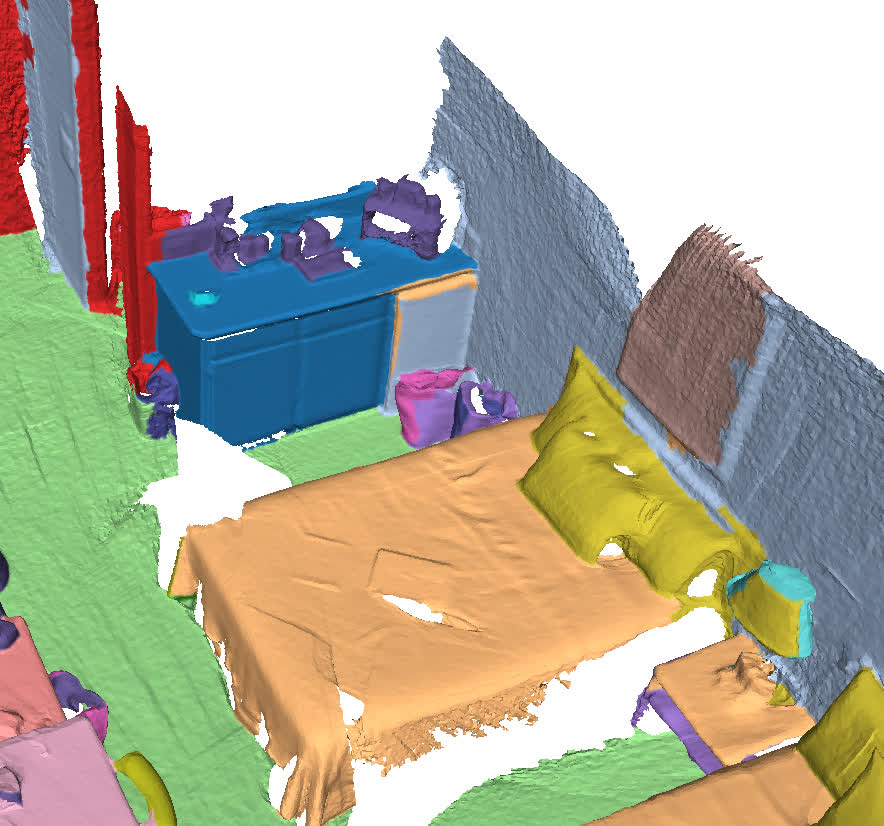}};
     \draw[red,ultra thick,rounded corners] (1.2,1.2) rectangle (2.3,2.5);
    \draw[red,ultra thick,rounded corners] (2.4,.1) rectangle (3.3,1.2);
    \end{tikzpicture}
    & 
     \begin{tikzpicture}
    \node[anchor=south west,inner sep=0] (image) at (0,0) { \includegraphics[height=\sz\textwidth]{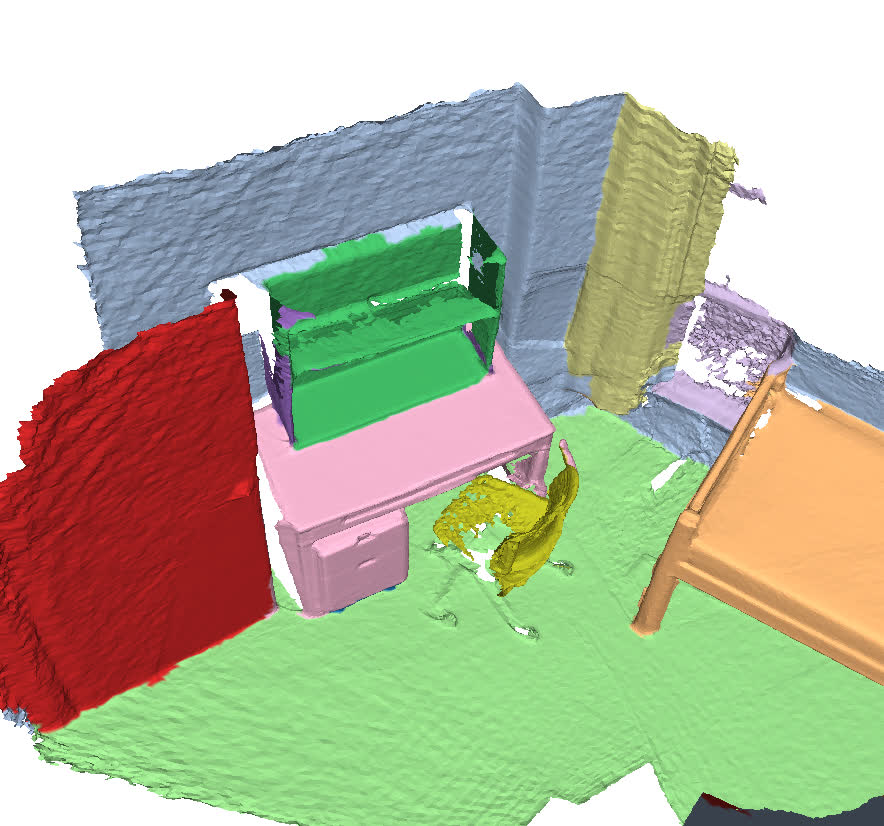}};
    \draw[red,ultra thick,rounded corners] (0.9,.6) rectangle (1.9,1.7);
    \end{tikzpicture}
    \\
    \rotatebox{90}{\textbf{{~~~~~~~Expert Labels}}}
    \begin{tikzpicture}
    \node[anchor=south west,inner sep=0] (image) at (0,0) { \includegraphics[height=\sz\textwidth]{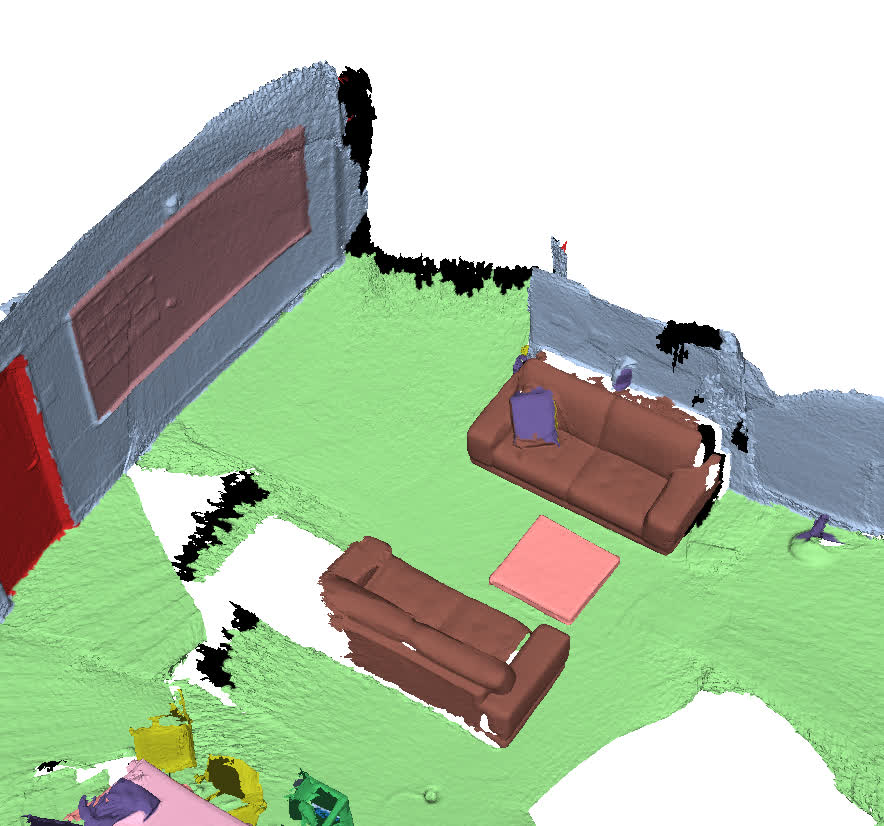}};
    \draw[red,ultra thick,rounded corners] (1.5,0.6) rectangle (2.8,2.);
    \end{tikzpicture} &
    \begin{tikzpicture}
    \node[anchor=south west,inner sep=0] (image) at (0,0) { \includegraphics[height=\sz\textwidth]{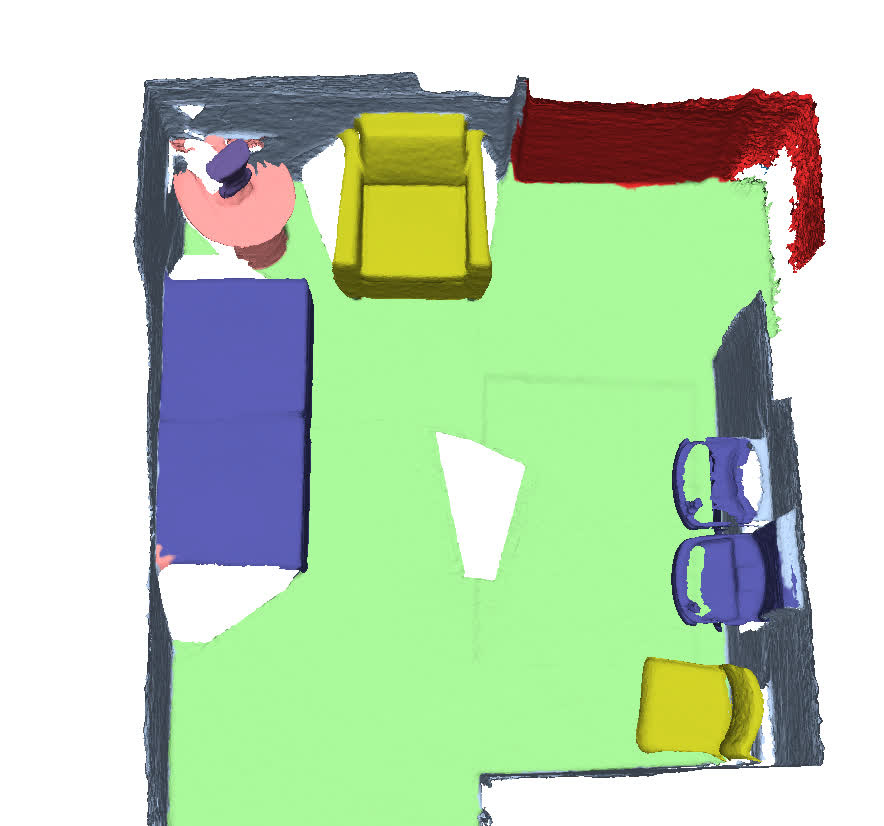}};
    \draw[red,ultra thick,rounded corners] (0.5,1.) rectangle (2.,3.);
    \end{tikzpicture}
    & 
    \begin{tikzpicture}
    \node[anchor=south west,inner sep=0] (image) at (0,0) { \includegraphics[height=\sz\textwidth]{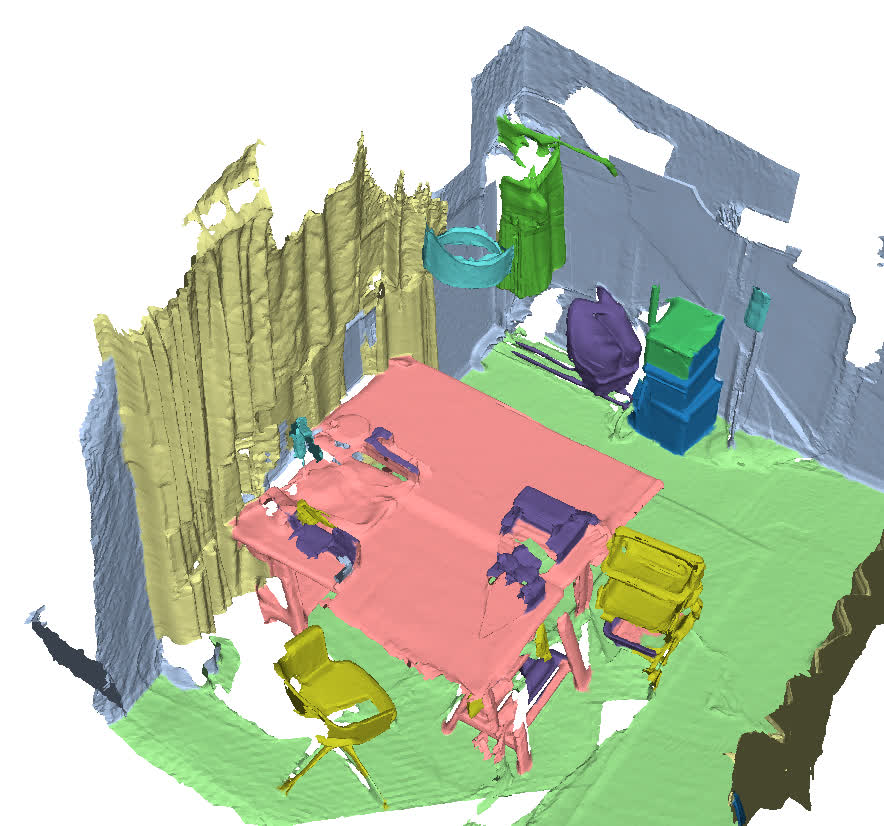}};
    \draw[red,ultra thick,rounded corners] (1.4,1.7) rectangle (2.4,2.9);
    \end{tikzpicture}
    & 
    \begin{tikzpicture}
    \node[anchor=south west,inner sep=0] (image) at (0,0) { \includegraphics[height=\sz\textwidth]{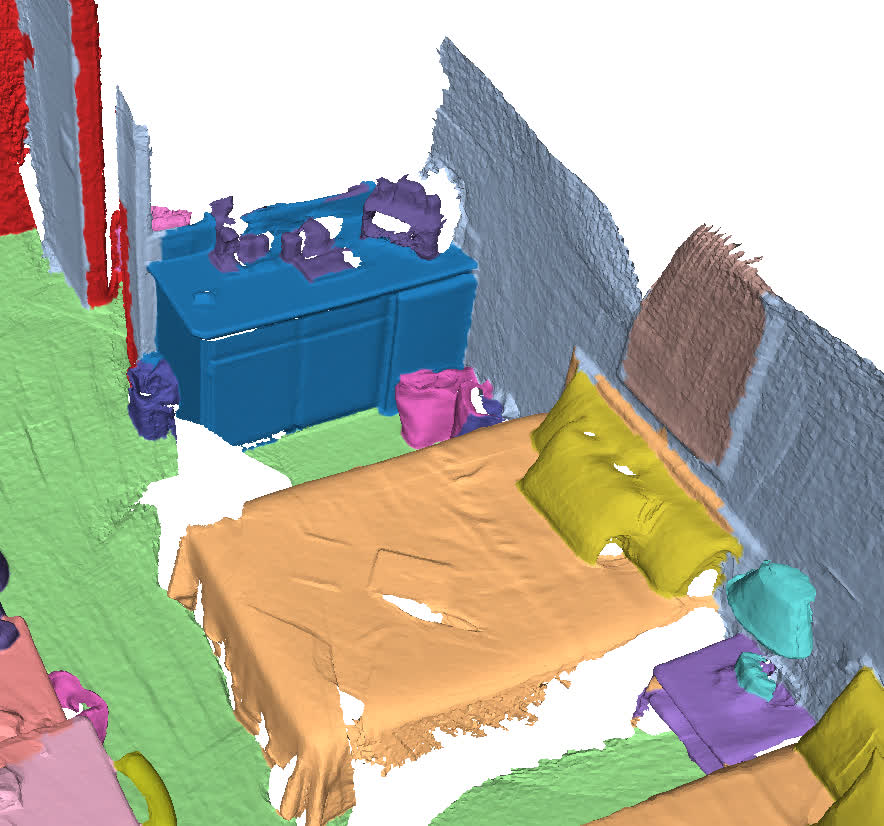}};
     \draw[red,ultra thick,rounded corners] (1.2,1.2) rectangle (2.3,2.5);
    \draw[red,ultra thick,rounded corners] (2.4,.1) rectangle (3.3,1.2);
    \end{tikzpicture}
    & 
    \begin{tikzpicture}
    \node[anchor=south west,inner sep=0] (image) at (0,0) { \includegraphics[ height=\sz\textwidth]{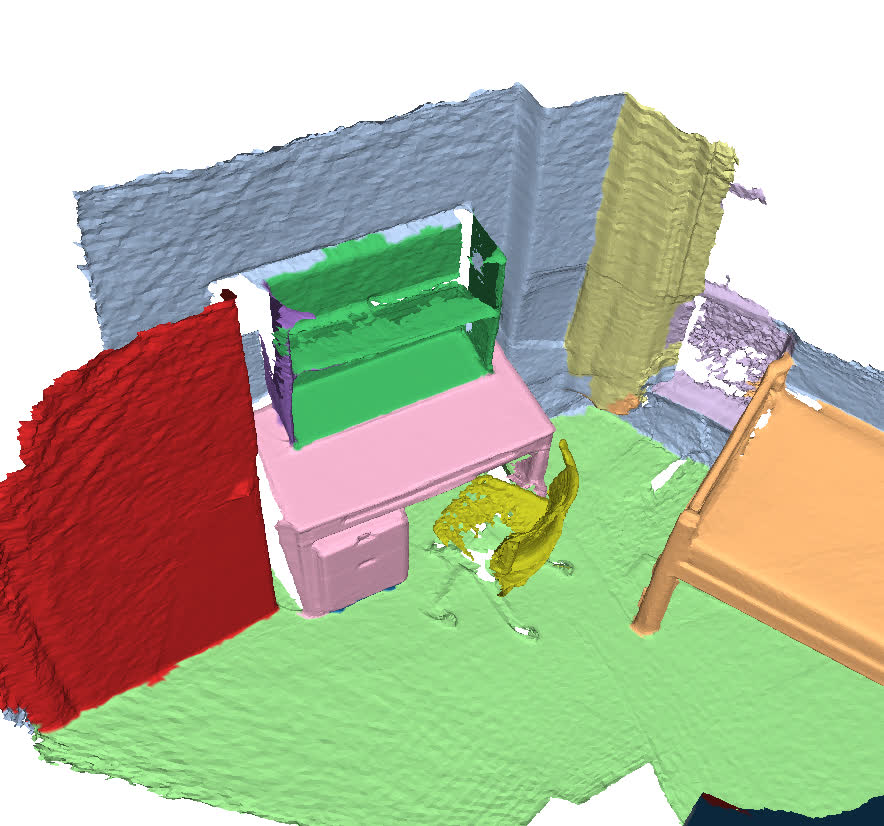}};
    \draw[red,ultra thick,rounded corners] (0.9,.6) rectangle (1.9,1.7);
    \end{tikzpicture}
    \\
    \rotatebox{90}{\textbf{{~~~~~~~~~~~Groundtruth}}}
    \begin{tikzpicture}
    \node[anchor=south west,inner sep=0] (image) at (0,0) { \includegraphics[height=\sz\textwidth]{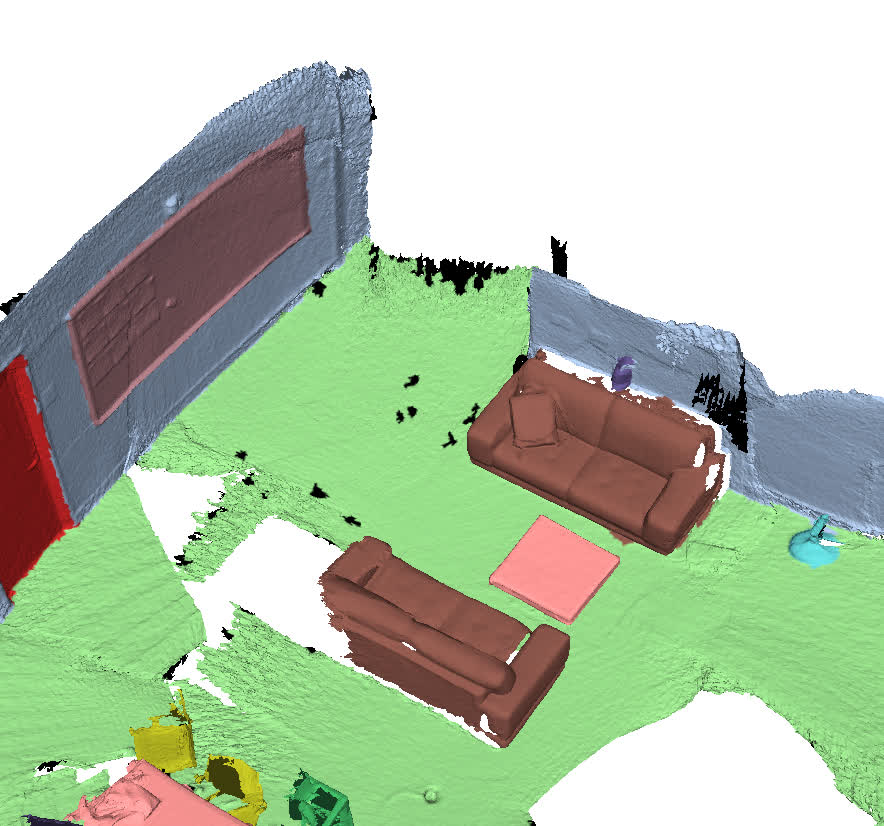}};
    \draw[red,ultra thick,rounded corners] (1.5,0.6) rectangle (2.8,2.);
    \end{tikzpicture}
    &

    \begin{tikzpicture}
    \node[anchor=south west,inner sep=0] (image) at (0,0) { \includegraphics[height=\sz\textwidth]{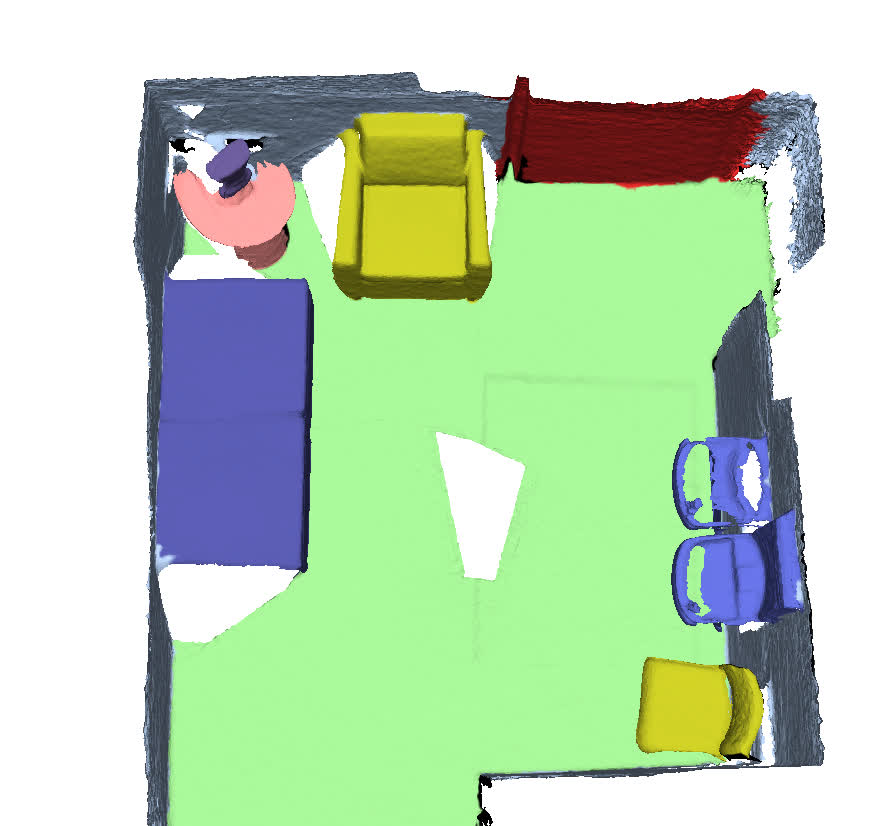}};
    \draw[red,ultra thick,rounded corners] (0.5,1.) rectangle (2.,3.);
    \end{tikzpicture}
    &
    \begin{tikzpicture}
    \node[anchor=south west,inner sep=0] (image) at (0,0) { \includegraphics[height=\sz\textwidth]{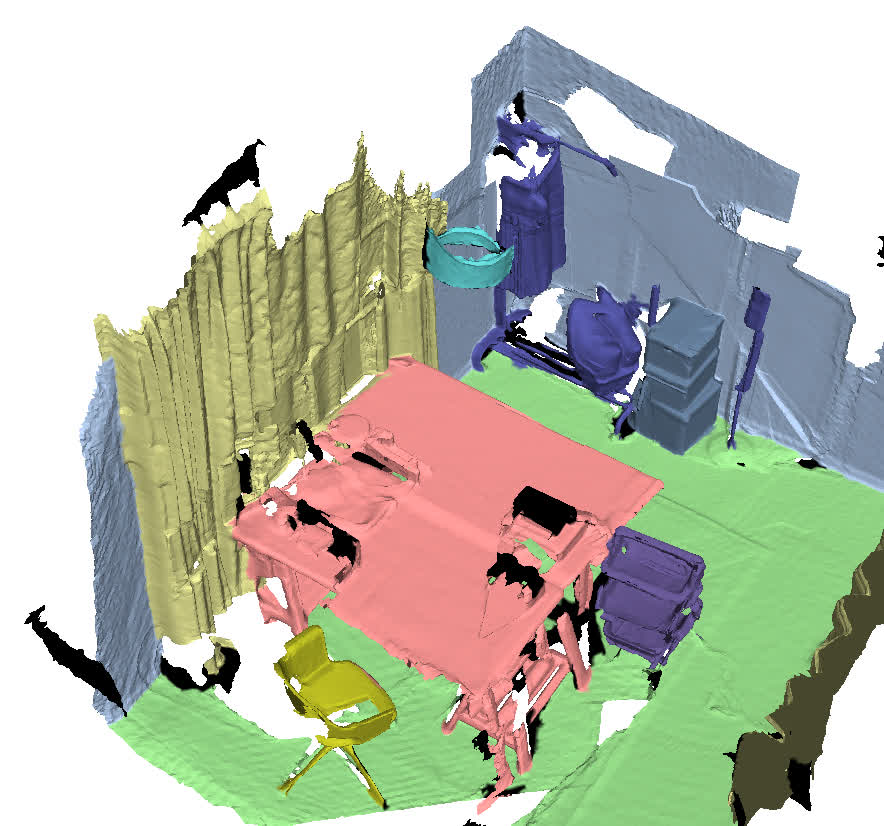}};
    \draw[red,ultra thick,rounded corners] (1.4,1.7) rectangle (2.4,2.9);
    \end{tikzpicture}
    &
    \begin{tikzpicture}
    \node[anchor=south west,inner sep=0] (image) at (0,0) { \includegraphics[height=\sz\textwidth]{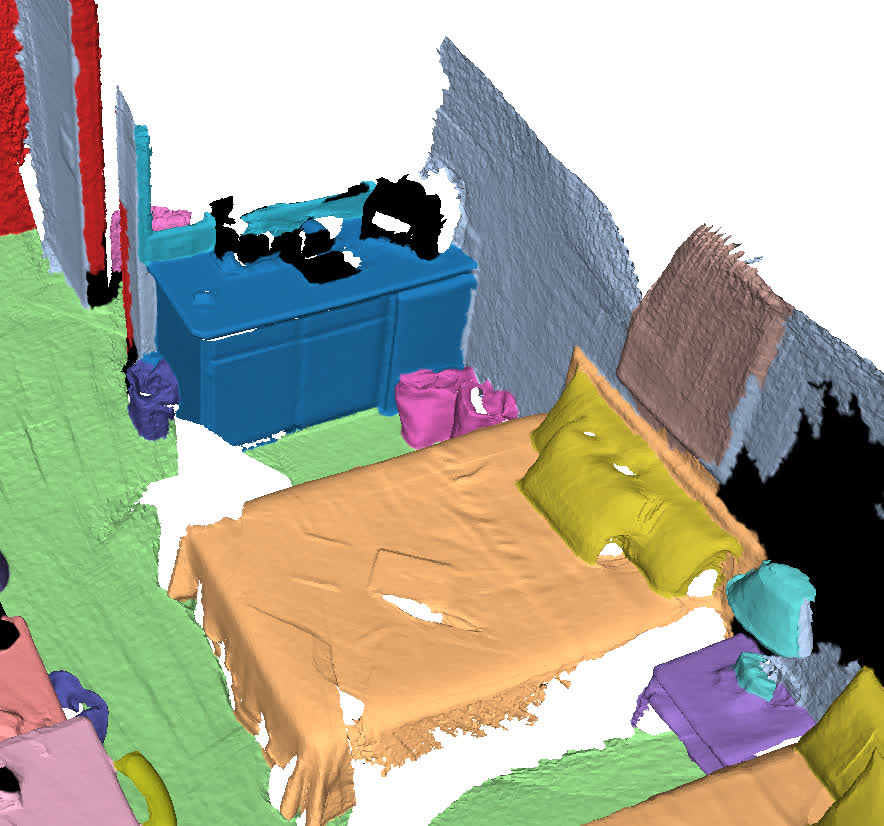}};
     \draw[red,ultra thick,rounded corners] (1.2,1.2) rectangle (2.3,2.5);
    \draw[red,ultra thick,rounded corners] (2.4,.1) rectangle (3.3,1.2);
    \end{tikzpicture}
    &    
    \begin{tikzpicture}
    \node[anchor=south west,inner sep=0] (image) at (0,0) { \includegraphics[height=\sz\textwidth]{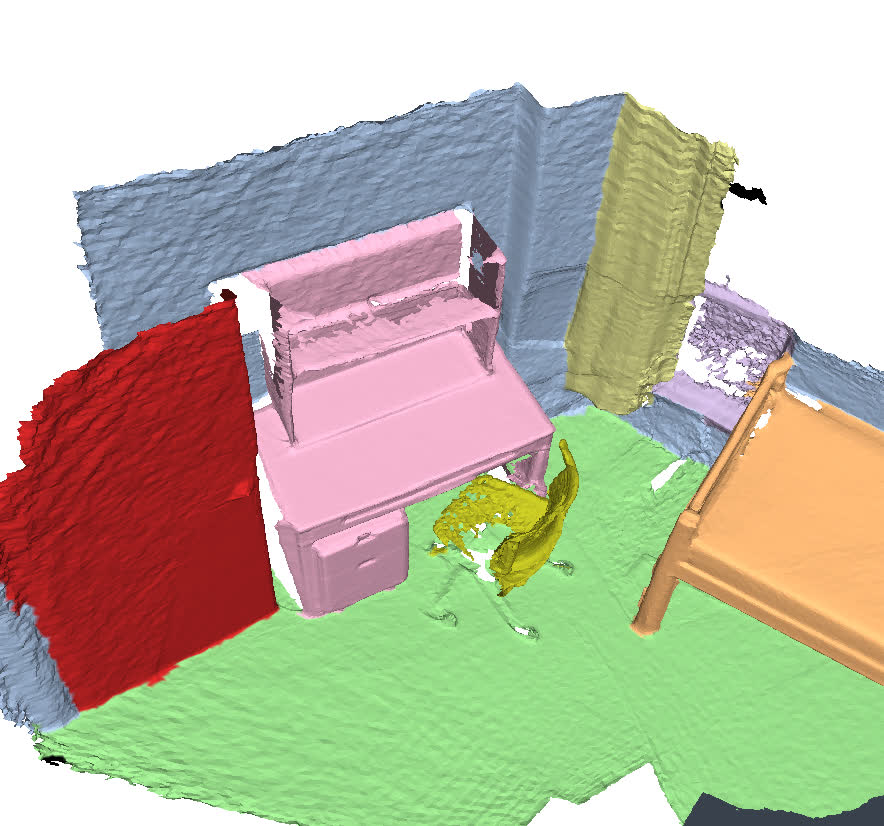}};
    \draw[red,ultra thick,rounded corners] (0.9,.6) rectangle (1.9,1.7);
    \end{tikzpicture}
    \\
\end{tabular}
\caption{
\textbf{Qualitative results of our proposed method and comparison between the different stages.}
The expert successfully selects the correct information from the two encoders. 
The 2D encoder is focused on object-level decisions
(\eg, table in column 1, bench in column 2) while the 3D information is used for fine details (\eg lamp in column 3, desk in column 5).
}
\label{fig:quali}
\end{figure*}

%% file: tables/per_class_expert_2d_3d.tex

\begin{table*}[t]
\vspace{5px}
  \setlength{\tabcolsep}{3px}%
  \centering
  \resizebox{\textwidth}{!}{
  \begin{tabular}{l c c c c c c c c c c c c c c c c c c c c c}
    \textbf{Method} & \textbf{mIoU}$\uparrow$ & \textbf{\rotatebox{70}{Wall}} & \textbf{\rotatebox{70}{Floor}} & \textbf{\rotatebox{70}{Cabinet}} & \textbf{\rotatebox{70}{Bed}} & \textbf{\rotatebox{70}{Chair}} & \textbf{\rotatebox{70}{Sofa}} & \textbf{\rotatebox{70}{Table}} &	\textbf{\rotatebox{70}{Door}} &	\textbf{\rotatebox{70}{Window}} &	\textbf{\rotatebox{70}{Bookshelf}} &	\textbf{\rotatebox{70}{Picture}} &	\textbf{\rotatebox{70}{Counter}} &	\textbf{\rotatebox{70}{Desk}} & 	\textbf{\rotatebox{70}{Curtain}} & 	\textbf{\rotatebox{70}{Fridge}} &\textbf{\rotatebox{70}{Shower Curt.}} &	\textbf{\rotatebox{70}{Toilet}} &	\rotatebox{70}{\textbf{Sink}} &	\rotatebox{70}{\textbf{Bathtub}} & \textbf{\rotatebox{70}{Other Furn.}}  \\
    \midrule
Ours - 2D	&	68.2	&	83.4	&	96.3	&	\textbf{60.1}	&	\underline{75.0}	&	\underline{85.8}	&	\textbf{77.0}	&	68.7	&	65.7	&	60.3	&	\textbf{66.7}	&	28.2	&	64.8	&	\underline{56.1}	&	68.3	&	\underline{61.8}	&	58.8	&	86.6	&	62.7	&	86.0	&	52.1	\\
Ours - 3D	&	\underline{69.0}	&	\underline{85.0}	&	96.8	&	59.2	&	73.9	&	\textbf{87.3}	&	\underline{75.8}	&	\underline{69.7}	&	\textbf{69.5}	&	\underline{61.3}	&	62.9	&	\underline{33.1}	&	\underline{65.7}	&	\underline{56.1}	&	\underline{68.4}	&	61.2	&	\underline{60.2}	&	\underline{89.0}	&	\underline{64.1}	&	\underline{87.0}	&	\underline{52.8}	\\
Ours - Temp	&	\textbf{70.6}	&	\textbf{85.6}	&	\textbf{96.9}	&	\underline{59.6}	&	\textbf{77.1}	&	\textbf{87.3}	&	75.0	&	\textbf{71.9}	&	\underline{68.4}	&	\textbf{65.8}	&	\underline{65.9}	&	\textbf{36.6}	&	\textbf{67.2}	&	\textbf{61.1}	&	\textbf{69.4}	&	\textbf{62.6}	&	\textbf{65.0}	&	\textbf{89.8}	&	\textbf{65.2}	&	\textbf{87.1}	&	\textbf{53.8}	\\
\midrule
$D_{F}$ = 64	&	67.9	&	\underline{84.4}	&	\underline{96.2}	&	59.4	&	76.9	&	85.6	&	73.0	&	69.5	&	\underline{66.3}	&	63.1	&	36.4	&	24.8	&	\underline{65.6}	&	\textbf{61.6}	&	\textbf{71.3}	&	\underline{61.5}	&	\textbf{67.1}	&	89.1	&	64.7	&	\textbf{89.4}	&	52.9	\\
$D_{F}$ = 40	&	\textbf{70.6}	&	\textbf{85.6}	&	\textbf{96.9}	&	\underline{59.6}	&	77.1	&	\textbf{87.3}	&	75.0	&	\textbf{71.9}	&	\textbf{68.4}	&	\textbf{65.8}	&	\textbf{65.9}	&	\textbf{36.6}	&	\textbf{67.2}	&	\underline{61.1}	&	69.4	&	\textbf{62.6}	&	\underline{65.0}	&	89.8	&	\underline{65.2}	&	87.1	&	\underline{53.8}	\\
$D_{F}$ = 32	&	\underline{68.2}	&	83.0	&	94.9	&	\textbf{60.8}	&	\textbf{78.5}	&	\underline{86.3}	&	\textbf{76.0}	&	\underline{69.8}	&	61.6	&	58.7	&	\underline{63.6}	&	24.6	&	64.5	&	\underline{61.1}	&	\underline{70.0}	&	60.9	&	50.4	&	\underline{90.6}	&	\textbf{67.5}	&	\underline{87.5}	&	53.0	\\
$D_{F}$ = 16	&	68.1	&	83.2	&	95.2	&	58.5	&	\underline{77.9}	&	84.9	&	\underline{75.8}	&	69.4	&	61.5	&	\underline{63.8}	&	52.9	&	\underline{33.0}	&	60.1	&	58.2	&	64.9	&	\underline{61.5}	&	64.7	&	\textbf{90.8}	&	63.9	&	85.9	&	\textbf{55.0}	\\
\midrule
Xception	&	\textbf{70.6}	&	\textbf{85.6}	&	\textbf{96.9}	&	\textbf{59.6}	&	\textbf{77.1}	&	\textbf{87.3}	&	\textbf{75.0}	&	\textbf{71.9}	&	\textbf{68.4}	&	\textbf{65.8}	&	\textbf{65.9}	&	36.6	&	\textbf{67.2}	&	61.1	&	\textbf{69.4}	&	\textbf{62.6}	&	\textbf{65.0}	&	\textbf{89.8}	&	\textbf{65.2}	&	\textbf{87.1}	&	\textbf{53.8}	\\
MobileNet	&	66.0	&	81.2	&	95.9	&	57.8	&	73.9	&	83.1	&	70.8	&	68.0	&	60.3	&	50.2	&	63.9	&	\textbf{37.3}	&	63.1	&	\textbf{61.9}	&	63.4	&	56.3	&	50.1	&	88.6	&	58.5	&	86.2	&	50.0	\\ 
    \bottomrule
  \end{tabular}
  }
  \vspace{-5px}
  \caption{\textbf{Ablating different aspects of our pipeline on ScanNet~\cite{Dai-et-al-CVPR-2017} validation set.} We show that the expert selects valuable information from the two encoders and the existing scene representation by evaluating the individual pipeline outputs. The expert network consistently improves upon the two other stages in terms of IoU.
  We also evaluate the impact of the stored feature dimension $D_F$ on the overall result. 
  While the smaller feature sizes suffer from compression due to limited capacity the larger features ($D_F = 64$) suffer from slight overfitting. Finally, we compare the Xception 2D encoder to the lighter MobileNet ($\times 20$ fewer parameters). 
  Unsurprisingly, the smaller encoder leads to a slight deterioration of performance, but the results indicate potential for runtime-accuracy tradeoffs in time-critical applications.
  }
  \label{tab:ablation_2d_3d_expert} 
\end{table*}

%% file: tables/semantic_scannet_scenenn.tex


\begin{table}
  \setlength{\tabcolsep}{3px}%
  \centering
  \resizebox{\columnwidth}{!}{
  \begin{tabular}{l c c c c c c}
    \toprule
    \multicolumn{6}{c}{\textbf{3D Semantic Segmentation}}  \\ 
    \midrule
    & & & \multicolumn{1}{c}{ScanNet} & \multicolumn{2}{c}{SceneNN}\\
    \cmidrule(r){4-4} \cmidrule(r){5-6}
    & Processing & Res. [cm] & val.\,mIoU & wIoU & mAcc \\
    \midrule
    SemanticFusion~\cite{Mccormac-et-al-ICRA-2017} & \textcolor{Orange}{Global} & N/A & 42.3 & \textbf{47.1} & \underline{58.5} \\
    PanopticFusion~\cite{narita2019panopticfusion} & \textcolor{Orange}{Global} & 2.4 & \underline{53.1} & -- & -- \\
    InsConv~\cite{liu2022ins} & \textcolor{Orange}{Global} & 2 & \textbf{72.4} & -- & \textbf{79.5} \\
    \midrule
    SemanticReconstruction~\cite{jeon2018semantic} & \textcolor{Green}{Local} & N/A  & 44.0 & -- & -- \\
    ProgressiveFusion~\cite{pham2019progressivefusion} & \textcolor{Green}{Local} & 0.8 & 55.0 & 52.2 & 61.6 \\
    FusionAware~\cite{zhang2020fusion} & \textcolor{Green}{Local} & N/A & 67.2 & 63.9 & 71.7 \\
    SVCNN~\cite{huang2021supervoxel} & \textcolor{Green}{Local} & N/A & \underline{68.3} & \textbf{69.0} & \textbf{76.9} \\
    \textbf{\name{} (Ours)} & \textcolor{Green}{Local} & 4 & \textbf{70.6} & \underline{67.8} & \textbf{76.9} \\
    \bottomrule
  \end{tabular}
  }
  \vspace{0px}
  \caption{\textbf{3D Semantic Segmentation on ScanNet and SceneNN.}
  Scores are mean intersection over union (mIoU) on ScanNet~\cite{Dai-et-al-CVPR-2017} validation, the mean accuracy (mAcc) and weighted IoU (wIoU) on SceneNN \cite{pham2019progressivefusion}.
  All other scores are as reported in \cite{huang2021supervoxel} and \cite{liu2022ins}.
  \vspace{-10px}}
  \label{tab:semantic_scannet_scenenn}
\end{table}

%% file: sec/05_conclusion.tex
\section{Conclusion}
We presented a novel pipeline for online joint geometric and semantic 3D reconstruction. 
The pipeline consists of three components and a learned scene representation that represents the scene as sparse voxel grid.
In order to leverage the complementary nature of 2D and 3D information, the first two stages encode 2D RGB-D data and enhance the encoded features with 3D spatial information.
At the heart of your pipeline sits a temporal expert fusion network, that sequentially updates the learned scene representation. 
This network attends to the 2D, 3D, and existing features to extract relevant information for the updates.
We experimentally show that this design improves the performance on semantic segmentation upon the two individual branches.

\parag{Acknowledgments.}
This research is partially supported by Toshiba, an ETH AI Center PostDoc Fellowship, and an ETH Career Seed Award funded through the ETH Zurich Foundation.

%% file: sec/06_appendix.tex
\appendix

\section{More Pipeline Details}
\label{sec:pipeline}
In this part, we present additional information about our proposed pipeline. 

\parag{3D model architecture.}
The UNet consists of four blocks and each block consists of two residual layers with interleaved batch norm and ReLU layers. 
The output of this layer is downsampled to the resolution of the next block using a convolutional layer with a kernel size and stride of two. 
The output of the lower block is processed by a transposed convolution and combined with the skip connection of the current level.
Similar to the 2D encoder, the 3D encoder is also supervised using an auxiliary head for semantic segmentation enforcing an as good performance as possible from a single 3D frame.

\parag{Auxiliary heads.}
After each sub-network, we leverage a additional classification head to supervise the pipeline using an auxiliary semantic loss. 
The additional head consumes the features predicted by the network and passes them through a two-layer MLP that is interleaved with batch normalization and ReLU activation.
For all sub-networks, we use the exact same architecture in order to uniformly constrain the feature space. 
The goal of these additional heads is to enable an auxiliary loss that enforces the sub-networks to encode as much semantic information as they possibly can extract from the given input. 

\parag{Positional encoding of encoder features.}
To enable the spatio-temporal expert network to make optimal fusion decisions, it has to learn from which encoder to select information from. 
However, transformer architectures are invariant to permutations in their input sequence. 
This means that the network cannot learn from which source a specific feature in the input is obtained. 
Thus, we have to encode the source information into the feature that is passed to the network. 
As we keep the ordering of the different sources fixed in the input sequence, we can treat the source information like positional information. 
Therefore, we encode the source information using positional encoding as it is standardized in transformer networks.
This positional encoding allows to inform the temporal expert network from which source a specific feature is coming.

\parag{Avoiding feature drift.}
When using learned features during iterative fusion of sensor measurements, the problem of feature drift can arise. 
Feature drift is a problem during inference. 
During inference, features might drift outside the distribution they have been trained on.
Thus, the trained neural networks fail to handle them properly as they represent  unknown information. 
Oftentimes, this drift manifests itself by the growing scale of the norm of the individual features.
In our pipeline, this is primarily caused by the bias of the different layers in the 3D encoder and the spatio-temporal expert network that are iteratively added on top of the features.
Thus, we have to prevent that during training and testing. 

While other works, \eg{}, INS-Conv~\cite{liu2022ins}, address this issue by performing network passes over the entire scene (which can limit scalability to large-scale scenes), we approach this problem such that it is more suitable for online fusion:
Given the observation that the growing norm is the main driver of feature drift, we utilize a simple normalization strategy. 
To this end, after all sub-networks -- 2D encoder, 3D encoder, and spatio-temporal expert network -- we pass the final features through a one-dimensional layer norm. 
This yields normalized features and prevents their norm from growing outside the training distribution.

\section{More Results}
\label{sec:results}

\subsection{Results on limited data}
\begin{figure}
    \centering
    \includegraphics[width=\columnwidth]{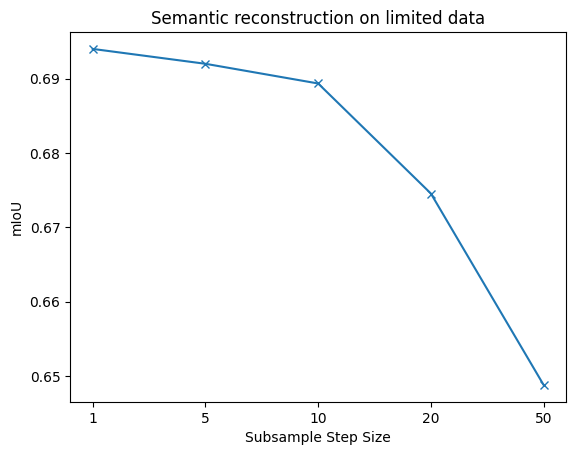}
    \caption{\textbf{Our model on limited data.} We evaluate the performance of the proposed pipeline on limited data by subsampling the trajectories with different step sizes. We can see that up to a subsampling step size of 10, the performance drop is insignificant while above that it starts to drop. However, also under these conditions the performance does not completely deteriorate.}
    \label{fig:limited-data}
\end{figure}

In order to evaluate the performance of our method on limited data, we randomly selected 10 scenes from the ScanNet validation set. 
Then, we subsampled the trajectories with different set sizes in order to simulate limited data.
In Figure~\ref{fig:limited-data}, we show the performance for these different step sizes.
For small step sizes, the performance only slighlty changes. 
For larger step sizes (step size $>$ 10), the performance starts to drop to a subpar performance.
However, even for a very large step size (step size = 50), the performance does not completely deteriorate.

\subsection{More qualitative results}

In Figure~\ref{fig:quali-supp-mat-1} and~\ref{fig:quali-supp-mat-2}, we show additional qualitative results emphasizing the benefits of leveraging the complementary information obtained from 2D and 3D. 
As it can be seen from comparing the outputs of the 2D and 3D networks, the 2D network is mainly responsible for avoiding object-scale confusion of the semantic class. 
\Eg, a desk that this erroneously confused as a table by the 3D network is correctly classified by 2D network. 
The spatio-temporal expert network learns to select the correct information and corrects the errors. 
On the other hand, the 3D network mainly refines low-level misclassifications of points in the scene and acts as a neural regularizer making sure that all points belonging to the same object are consistently labelled. 
\Eg, this can be seen in the case of the chair or desks, where only partially wrong predictions made by the 2D network, which are subsequently refined by the 3D network.
The spatio-temporal expert learns to leverage this information from the 3D network. 
In summary, the output of the spatio-temporal expert network is consistently better than the output of the two separate branches by fusing their complementary information with the already integrated information.
\input{3dv/figures/quali_supp_mat}



\subsection{Additional discussion of results}

\begin{figure*}
    \centering
    \includegraphics[width=\textwidth]{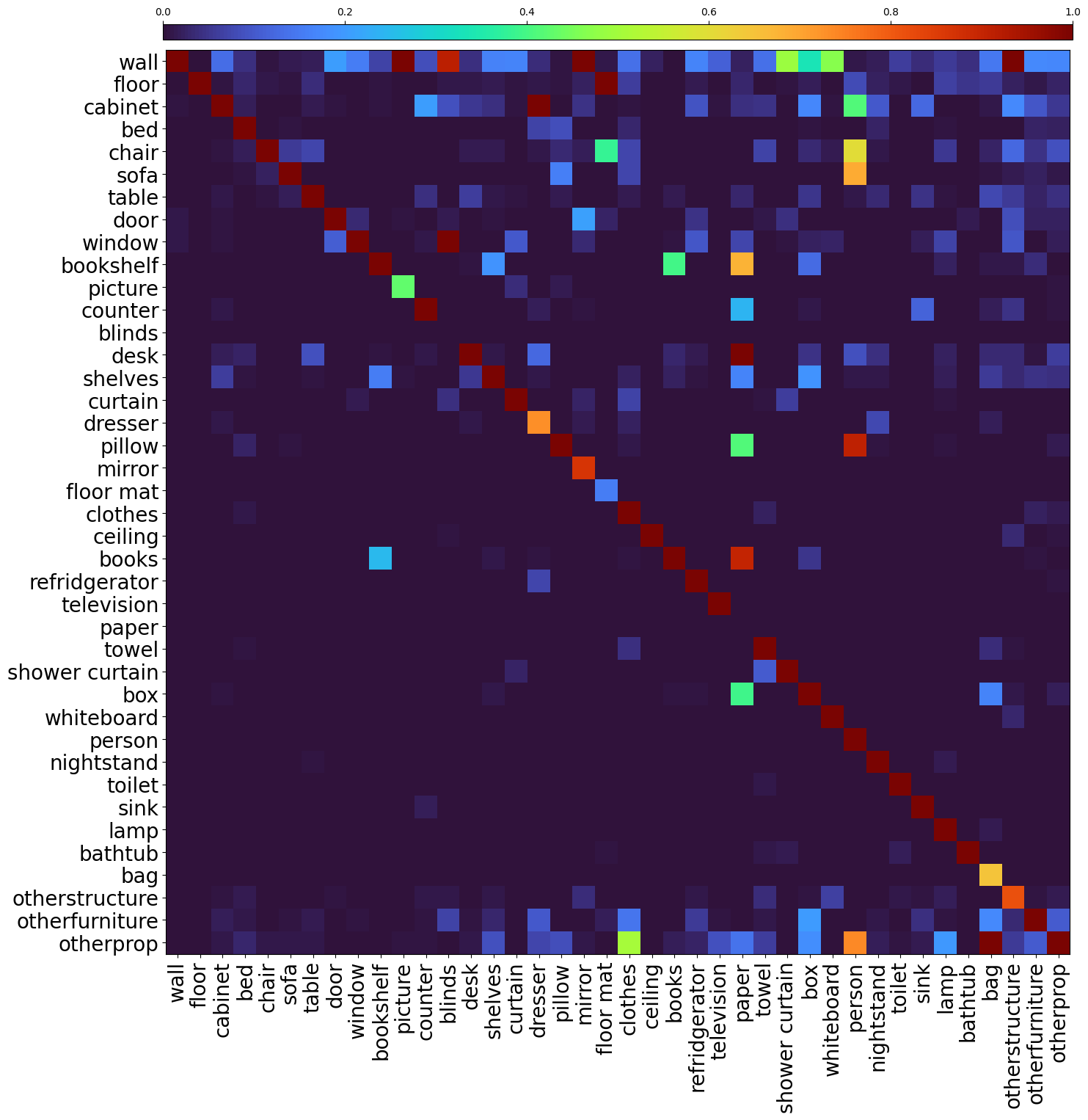}
    \caption{\textbf{Confusion matrix of ScanNet validation set results.}}
    \label{fig:confusion_matrix}
\end{figure*}

In Figure~\ref{fig:confusion_matrix}, we visualize the confusion matrix for our pipeline on the ScanNetv2 dataset. 
To better visualize the differences, the values are normalized in each column, which is the relevant axis for evaluation.

A difficult class on ScanNetv2 is the `picture' category. 
One can see from the confusion matrix that this class is oftentimes confused for the wall class.
However, this is also a class where the complementary benefit of 2D and 3D information is striking. 
We can show a significant improvement over the two encoders as well as to all baselines.

We also address the problems with the two classes where a significant drop in performance is visible on the test set.
We first focus on the bookshelf class. 
There are two issues with the bookshelf class. 
The first issue is caused by the hierarchical nature of the labels. 
\Ie, books inside a bookshelf are oftentimes labelled as bookshelf. 
However, our method predominantly ``correctly`` classifies these books as books.
This causes a significant drop in performance.
The second issue is caused by the confusion between bookshelf and shelf. 
These are not consistently labelled in the dataset.
Thus, our pipeline does not robustly learn this distinction and miss-classifies this category.
The second class are refrigerators. 
For this class, the confusion is fuzzier.
\Ie, refrigerators are confused with different classes (wall, other furniture, window, and wall).

\subsection{Discussion of baseline comparison}

Firstly, we would like to further detail the differences to the main baselines.
While our pipeline purely operates in a local frame, some other baselines aggregate global information at some stage in the pipeline.
\Eg, INS-Conv~\cite{liu2022ins} runs a network pass over the entire reconstruction of the scene in order to avoid feature drift. 
This is expensive when the pipeline is applied in the reconstruction of large-scale scenes. 
There it can become a severe bottleneck to run global processing to avoid feature drift.
We improve this issue by carefully designing the pipeline with normalization layers in order to avoid this feature drift without requiring global processing.

Secondly, due to the difficulty of unavailable code for the global methods, we cannot get more qualitative and quantitative results than the information in their papers.
Thus, in order to further facilitate research in this field, we will publish the source code on publication of the paper.

\subsection{Why is there no qualitative comparison to baselines?}
As the reader might have noted, there is no qualitative comparison to one of the baselines. 
This is due to the fact that it is not standard practice to publish the full pipeline source code for the proposed methods\footnote{INS-Conv~\cite{liu2022ins} has released a code snippet on \href{https://github.com/THU-luvision/INS-Conv}{https://github.com/THU-luvision/INS-Conv} for their proposed model architecture, but not the full training and evaluation pipeline. SVCNN~\cite{huang2021supervoxel} has released the training code for their neural network, but not the full model on \href{https://github.com/shishenghuang/SVNet_jittor}{https://github.com/shishenghuang/SVNet\_jittor}}
Thus, it is not possible to show qualitative results on reconstructed data in a baseline comparison. 
With this paper, we aim to change this practice by releasing the source code on acceptance of this manuscript in order to foster more research in this field.

%% file: figures/quali_supp_mat.tex
\begin{figure*}[ht]
\centering
\footnotesize
\renewcommand{\arraystretch}{0.5} 
\begin{tabular}{cccc}
    \textbf{2D Output} & \textbf{3D Output} & \textbf{Expert Output} & \textbf{Ground-truth} \\
    \includegraphics[width=4cm, height=4cm]{3dv/figures/qualitative_semantic_scannet_supp_mat/1_2d.png} & 
    \includegraphics[width=4cm, height=4cm]{3dv/figures/qualitative_semantic_scannet_supp_mat/1_3d.png} & 
    \includegraphics[width=4cm, height=4cm]{3dv/figures/qualitative_semantic_scannet_supp_mat/1_expert.png} & 
    \includegraphics[width=4cm, height=4cm]{3dv/figures/qualitative_semantic_scannet_supp_mat/1_expert.png} \\
    \\
    \includegraphics[width=4cm, height=4cm]{3dv/figures/qualitative_semantic_scannet_supp_mat/2_2d.png} & 
    \includegraphics[width=4cm, height=4cm]{3dv/figures/qualitative_semantic_scannet_supp_mat/2_3d.png} & 
    \includegraphics[width=4cm, height=4cm]{3dv/figures/qualitative_semantic_scannet_supp_mat/2_expert.png} & 
    \includegraphics[width=4cm, height=4cm]{3dv/figures/qualitative_semantic_scannet_supp_mat/2_gt.png} \\
    \\
    \includegraphics[width=4cm, height=4cm]{3dv/figures/qualitative_semantic_scannet_supp_mat/3_2d.png} & 
    \includegraphics[width=4cm, height=4cm]{3dv/figures/qualitative_semantic_scannet_supp_mat/3_3d.png} & 
    \includegraphics[width=4cm, height=4cm]{3dv/figures/qualitative_semantic_scannet_supp_mat/3_expert.png} & 
    \includegraphics[width=4cm, height=4cm]{3dv/figures/qualitative_semantic_scannet_supp_mat/3_gt.png} \\
    \\
    \includegraphics[width=4cm, height=4cm]{3dv/figures/qualitative_semantic_scannet_supp_mat/4_2d.png} & 
    \includegraphics[width=4cm, height=4cm]{3dv/figures/qualitative_semantic_scannet_supp_mat/4_3d.png} & 
    \includegraphics[width=4cm, height=4cm]{3dv/figures/qualitative_semantic_scannet_supp_mat/4_expert.png} & 
    \includegraphics[width=4cm, height=4cm]{3dv/figures/qualitative_semantic_scannet_supp_mat/4_gt.png} \\
    \\
    \includegraphics[width=4cm, height=4cm]{3dv/figures/qualitative_semantic_scannet_supp_mat/5_2d.png} & 
    \includegraphics[width=4cm, height=4cm]{3dv/figures/qualitative_semantic_scannet_supp_mat/5_3d.png} & 
    \includegraphics[width=4cm, height=4cm]{3dv/figures/qualitative_semantic_scannet_supp_mat/5_expert.png} & 
    \includegraphics[width=4cm, height=4cm]{3dv/figures/qualitative_semantic_scannet_supp_mat/5_gt.png} \\
    \\
    \textbf{2D Output} & \textbf{3D Output} & \textbf{Expert Output} & \textbf{Ground-truth}

\end{tabular}
\caption{\textbf{Additional qualitative results on the ScanNetv2 validation set.} }
\label{fig:quali-supp-mat-1}
\end{figure*}

\begin{figure*}[ht]
\centering
\footnotesize
\renewcommand{\arraystretch}{0.5} 
\begin{tabular}{cccc}
    \textbf{2D Output} & \textbf{3D Output} & \textbf{Expert Output} & \textbf{Ground-truth} \\
    \includegraphics[width=4cm, height=4cm]{3dv/figures/qualitative_semantic_scannet_supp_mat/6_2d.png} & 
    \includegraphics[width=4cm, height=4cm]{3dv/figures/qualitative_semantic_scannet_supp_mat/6_3d.png} & 
    \includegraphics[width=4cm, height=4cm]{3dv/figures/qualitative_semantic_scannet_supp_mat/6_expert.png} & 
    \includegraphics[width=4cm, height=4cm]{3dv/figures/qualitative_semantic_scannet_supp_mat/6_expert.png} \\
    \\
    \includegraphics[width=4cm, height=4cm]{3dv/figures/qualitative_semantic_scannet_supp_mat/7_2d.png} & 
    \includegraphics[width=4cm, height=4cm]{3dv/figures/qualitative_semantic_scannet_supp_mat/7_3d.png} & 
    \includegraphics[width=4cm, height=4cm]{3dv/figures/qualitative_semantic_scannet_supp_mat/7_expert.png} & 
    \includegraphics[width=4cm, height=4cm]{3dv/figures/qualitative_semantic_scannet_supp_mat/7_gt.png} \\
    \\
    \includegraphics[width=4cm, height=4cm]{3dv/figures/qualitative_semantic_scannet_supp_mat/8_2d.png} & 
    \includegraphics[width=4cm, height=4cm]{3dv/figures/qualitative_semantic_scannet_supp_mat/8_3d.png} & 
    \includegraphics[width=4cm, height=4cm]{3dv/figures/qualitative_semantic_scannet_supp_mat/8_expert.png} & 
    \includegraphics[width=4cm, height=4cm]{3dv/figures/qualitative_semantic_scannet_supp_mat/8_gt.png} \\
    \\
    \includegraphics[width=4cm, height=4cm]{3dv/figures/qualitative_semantic_scannet_supp_mat/9_2d.png} & 
    \includegraphics[width=4cm, height=4cm]{3dv/figures/qualitative_semantic_scannet_supp_mat/9_3d.png} & 
    \includegraphics[width=4cm, height=4cm]{3dv/figures/qualitative_semantic_scannet_supp_mat/9_expert.png} & 
    \includegraphics[width=4cm, height=4cm]{3dv/figures/qualitative_semantic_scannet_supp_mat/9_gt.png} \\
    \\
    \includegraphics[width=4cm, height=4cm]{3dv/figures/qualitative_semantic_scannet_supp_mat/10_2d.png} & 
    \includegraphics[width=4cm, height=4cm]{3dv/figures/qualitative_semantic_scannet_supp_mat/10_3d.png} & 
    \includegraphics[width=4cm, height=4cm]{3dv/figures/qualitative_semantic_scannet_supp_mat/10_expert.png} & 
    \includegraphics[width=4cm, height=4cm]{3dv/figures/qualitative_semantic_scannet_supp_mat/10_gt.png} \\
    \\
    \textbf{2D Output} & \textbf{3D Output} & \textbf{Expert Output} & \textbf{Ground-truth} \\

\end{tabular}
\caption{\textbf{Additional qualitative results on the ScanNetv2 validation set.}}
\label{fig:quali-supp-mat-2}
\end{figure*}